\documentclass[10pt,twocolumn,letterpaper]{article}

\usepackage[pagenumbers]{wacv} % To force page numbers, e.g. for an arXiv version

\usepackage[accsupp]{axessibility}  % Improves PDF readability for those with disabilities.
\usepackage{graphicx}
\usepackage{booktabs}
\usepackage{multirow}
\usepackage{tabularx}
\usepackage{wrapfig}
\usepackage[normalem]{ulem}
\useunder{\uline}{\ul}{}
\usepackage{bbm}
\usepackage{xspace}
\definecolor{lightblue}{HTML}{01feff}
\definecolor{darkerblue}{HTML}{807fff}
\definecolor{brightpurple}{HTML}{ff00ff}
\definecolor{imageMetric}{HTML}{4F587D}
\definecolor{camMetric}{HTML}{657F49}
\definecolor{3dMetric}{HTML}{C42536}
\newcommand{\name}{ViCoDR\xspace}
\newcommand{\norm}[1]{\left\lVert#1\right\rVert}

\definecolor{wacvblue}{rgb}{0.21,0.49,0.74}
\usepackage[pagebackref,breaklinks,colorlinks,allcolors=wacvblue]{hyperref}

\usepackage[capitalize]{cleveref}
\crefname{section}{Sec.}{Secs.}
\Crefname{section}{Section}{Sections}
\crefname{table}{Tab.}{Tabs.}
\Crefname{table}{Table}{Tables}
\def\wacvPaperID{292}
\def\confName{WACV}
\def\confYear{2027}

\title{Do View-Consistent Representations Matter for Camera-Controlled Video Generation?}
\author{Duolikun Danier\textsuperscript{1} \hspace{0.2em} Ge Gao\textsuperscript{2} \hspace{0.2em}  Chengxi Zeng\textsuperscript{2} \hspace{0.2em} Steven McDonagh\textsuperscript{1} \hspace{0.2em} Changjian Li\textsuperscript{1} \\ Hakan Bilen\textsuperscript{1} \hspace{0.2em} Oisin Mac Aodha\textsuperscript{1}  \\ 
\textsuperscript{1}University of Edinburgh \quad \textsuperscript{2} University of Bristol
}

\begin{document}
\maketitle

\begin{abstract}

Camera-controlled video generation has attracted increasing interest for its applications in simulation, gaming, and world modeling.
However, current models often deform rigid objects and scene structure as the viewpoint changes.
We investigate whether this failure is related to a measurable internal property: the view consistency of diffusion representations.
Across a diverse collection of camera-controlled image-to-video diffusion models, we find that models exhibiting more view-consistent representations generate videos with superior 3D consistency. 
We complement this correlation-based evidence with a within-model controlled intervention, \name, which directly supervises the \underline{\textbf{vi}}ew \underline{\textbf{co}}nsistency of internal \underline{\textbf{d}}iffusion \underline{\textbf{r}}epresentations, during training.
Across camera-controlled image-to-video, camera-controlled text-to-video, and multi-view generation tasks, \name consistently improves output 3D consistency.
Finally, we evaluate the loss used for these interventions as an auxiliary training objective by retraining a camera-controlled I2V model using either \name or existing methods for supervising diffusion representations under the same training setup.
We show that the \name loss improves output 3D consistency more reliably than the alternatives and remains effective across design variants, while exhibiting camera-control trade-offs and requiring additional training computation.
Together, these results establish view consistency of internal diffusion representations as an important property for analyzing and improving camera-controlled video generation.

\end{abstract}

\section{Introduction}\label{sec:intro}

Camera-controlled video generation has attracted increasing attention because it allows users and embodied agents to navigate generated scenes along prescribed camera trajectories, supporting applications in simulation~\cite{agarwal2025cosmos,yang2025instadrive}, gaming~\cite{liang2025diffusion,che2024gamegen}, and world modeling~\cite{team2025hunyuanworld,yu2026mosaicmem,li2025vmem,wu2026video}.
Advances in diffusion architectures~\cite{ho2020denoising,rombach2022high,peebles2023scalable}, training strategies~\cite{ma2024sit,chen2024diffusion}, and the availability of large-scale video data have substantially improved the realism of generated videos, making camera-controlled generation increasingly practical.
However, current models often fail to maintain 3D consistency~\cite{asim2025met3r,xie2025mvgbench}, meaning that their generated frames cannot be explained as projections of the same underlying 3D world.
This failure produces physically implausible outputs in which rigid objects and scene structures deform as the camera moves~\cite{liu2024syncdreamer,jeong2025track4gen} (see \cref{fig:i2v_qualitative}), degrading visual quality~\cite{zheng2025vbench}, hindering 3D reconstruction~\cite{schwarz2025generative,ren2025gen3c}, and reducing the immersiveness of interactive experiences~\cite{li2025vmem}.

To address this lack of 3D consistency, prior work has attempted to introduce additional 3D information obtained from off-the-shelf geometry estimation models to impose 3D-aware constraints. 
For example, some methods provide the generative model with extra inputs as conditions during training and inference, including re-projected point clouds~\cite{ren2025gen3c,team2025hunyuanworld}, point maps represented as 3D coordinates~\cite{gu2025diffusion}, or combinations thereof~\cite{cao2025mvgenmaster}.
However, these methods require external geometric signals during inference, increasing computation, and are prone to errors when the employed geometric estimators are inaccurate.
Other methods attempt to learn to jointly generate video pixels and the underlying geometry, using point maps~\cite{zhang2025world,szymanowicz2025bolt3d} or 3D Gaussians~\cite{schwarz2025generative}, but depend on paired high-quality 3D data for obtaining the training targets, which is expensive to collect at scale~\cite{wu2025cat4d}.

It has been shown that aligning diffusion representations with self-supervised vision encoders~\cite{oquab2024dinov}, known for their strong semantic features~\cite{chen2025probing}, can enhance \textit{image} generation quality~\cite{yu2025representation,leng2025repae}.
Inspired by this finding, recent works have extended such representation distillation to video generation~\cite{zhang2025videorepa,wu2025geometry} or have additionally supervised video diffusion representations on auxiliary tasks~\cite{jeong2025track4gen,huang2025jog3r}.
These objectives can improve generation without additional inference cost or ground-truth 3D supervision, but the representation-level factors associated with output 3D consistency remain poorly understood.

A promising candidate factor is multi-view consistency: visual representations in which features of the same 3D point remain consistent under view changes have been shown to encode geometric knowledge~\cite{yue2024improving} and benefit a range of tasks that require 3D understanding, including 3D reconstruction~\cite{michalkiewicz2025not}, pose estimation~\cite{ma2024imagenet3d}, depth estimation~\cite{yue2024improving}, and tracking~\cite{you2025multiview}.
This motivates the central question of our study: \emph{do multi-view-consistent diffusion representations matter for camera-controlled video generation?}
We address this via three complementary empirical questions.

\textbf{(Q1)} First, we ask whether internal view consistency is associated with output 3D consistency across camera-controlled video generation models.
We analyze models spanning different architectures, scales, and forms of geometric conditioning, and observe a strong cross-model relationship.
\textbf{(Q2)} Second, we ask whether deliberately enforcing view consistency in diffusion representations improves output 3D consistency across generation tasks.
We introduce a correspondence-based representation-level objective, \name (View-Consistent Diffusion Representations), as a controlled intervention and evaluate matched baseline/intervention pairs across camera-controlled I2V, camera-controlled T2V, and multi-view generation, providing evidence that view consistency in diffusion representations is important for output 3D consistency.
\textbf{(Q3)} Third, we study the comparative effectiveness and robustness of the \name loss used for these interventions as an auxiliary training objective.
Using CameraCtrl~\cite{he2025cameractrl} as a common testbed, we compare the \name loss with existing methods for supervising diffusion representations and analyze its formulation, pseudo-label dependence, loss-weight behavior, trade-offs, and computational cost.

To summarize, our contributions are:
\begin{itemize}
    \item A correlation-based analysis showing that view consistency of internal diffusion representations is strongly associated with output 3D consistency across diverse camera-controlled I2V models.
    \item A task-spanning intervention study showing that directly supervising view consistency in diffusion representations through \name, a correspondence-based objective, improves 3D consistency of generation across multiple models, including contemporary DiT-based models and those with explicit geometry conditioning.
    \item A focused CameraCtrl study evaluating the comparative effectiveness and robustness of the \name loss through comparisons with existing geometry-aware objectives and analysis of its formulation, pseudo-label dependence, loss-weight behavior, and trade-offs in 3D consistency, camera control, quality, and computation.
\end{itemize}

\section{Related Work}

\noindent{\bf Diffusion-based Video Generation.}
Early works on video generation~\cite{blattmann2023stable,he2022latent,ho2022video,guo2023animatediff,blattmann2023align,chen2023videocrafter1,wang2025lavie,zhang2024pia,xing2024dynamicrafter} extended image diffusion models to the temporal domain by training additional modules that capture inter-frame dependencies.
Empowered by large-scale data~\cite{chen2024panda,nan2024openvid} and transformer architectures~\cite{dosovitskiy2020vit,peebles2023scalable}, DiT-based video generators~\cite{brooks2024video,kong2024hunyuanvideo,yang2024cogvideox,wan2025wan,zheng2024open,chen2025goku} have shown impressive progress in generating high-resolution realistic videos.
In parallel, controllable video generation has expanded to conditioning on images~\cite{liu2024syncdreamer}, audio~\cite{he2024co}, human pose~\cite{hu2024animate}, motion~\cite{wang2024motionctrl}, and camera trajectories~\cite{he2025cameractrl,wang2024motionctrl,yang2024direct}.
Our study focuses on camera-controlled video generation, which allows users and embodied agents to navigate generated scenes along prescribed trajectories and supports novel view synthesis~\cite{mildenhall2021nerf,kerbl20233d,schwarz2025generative}, 3D scene generation~\cite{gao2024cat3d}, and interactive world models~\cite{li2025vmem,yu2026mosaicmem,wu2026video}.
As the same scene is observed from changing viewpoints, failures of 3D consistency become particularly apparent and consequential in this setting.

\noindent{\bf 3D-Consistent Camera Control.}
Camera trajectories can be specified implicitly through text descriptions~\cite{hu2024comd,li2025image} or explicitly through camera parameters~\cite{xu2024camco,he2025cameractrl,wang2024motionctrl,yang2024direct,kuang2024collaborative,bahmani2025ac3d,li2025realcam,bahmani2025vd3d}.
These camera-conditioning mechanisms have been used in both I2V and T2V models, while closely related multi-view generators synthesize views from specified camera viewpoints~\cite{liu2024syncdreamer,xie2025mvgbench}.
However, recent studies~\cite{asim2025met3r,xie2025mvgbench} have highlighted a persistent lack of 3D consistency in the outputs of these models.
Existing approaches to tackle the problem commonly provide additional 3D-consistent conditioning~\cite{hou2025trainingfree,ren2025gen3c,gu2025diffusion,cao2025mvgenmaster,yu2026mosaicmem} or jointly model pixels and geometry~\cite{zhang2025world,szymanowicz2025bolt3d,schwarz2025generative}.
The former rely on off-the-shelf geometry estimators during inference, while the latter are constrained by the availability of large-scale, high-quality 3D data.
Despite these advances, the relationship between internal diffusion representations and output 3D consistency remains unclear.

\noindent{\bf Evaluating 3D Consistency in Video Generation.}
Recent work has introduced metrics and benchmarks for consistency in generated views and worlds, including MEt3R~\cite{asim2025met3r}, MVGBench~\cite{xie2025mvgbench}, VBench~\cite{zheng2025vbench}, and WorldScore~\cite{duan2025worldscore}.
This enables systematic output-level comparisons across models.
We go beyond output-level evaluation by jointly measuring output 3D  and view consistency of diffusion representations across camera-controlled I2V models.

\noindent{\bf Diffusion Representations.}
A number of works have studied representations within diffusion models~\cite{man2024lexicon3d,zhan2024physd,danier2025depthcues,huang2025jog3r,bahmani2025ac3d,jeong2025track4gen}, revealing their utility across semantic and geometric tasks, \eg semantic correspondence~\cite{zhang2023tale}, visual grounding~\cite{tsagkas2024click}, 4D reconstruction~\cite{mai2025can}, and point tracking~\cite{jeong2025track4gen}.
More recently, several works~\cite{yu2025representation,leng2025repae} have studied the role of representations in generation itself and found that aligning diffusion representations with those of strong self-supervised vision encoders~\cite{oquab2024dinov} can enhance image generation.
Inspired by this, several video generation methods have utilized representation alignment or auxiliary supervision to promote temporal coherence, motion plausibility, or geometric structure~\cite{hwang2025cross,bhowmik2025moalign,zhang2025videorepa,wu2025geometry,huang2025jog3r,jeong2025track4gen}.
Among those targeting geometry, JOG3R~\cite{huang2025jog3r} and Geometry Forcing (GF)~\cite{wu2025geometry} supervise diffusion representations with VGGT~\cite{wang2025vggt} through point-map regression and feature distillation, respectively, whereas Track4Gen~\cite{jeong2025track4gen} provides point-tracking supervision.
However, the representation-level factors underlying the output-consistency gains of these objectives remain poorly understood.
In contrast, we study view consistency in diffusion representations as one such factor through cross-model correlation and matched intervention studies, and introduce a correspondence-based objective that directly promotes this property and improves output 3D consistency more than these alternatives.

\noindent{\bf View-Consistent Representations.}
View-consistent representations enable the establishment of geometric correspondence across views, which is essential for 3D understanding and reconstruction~\cite{rocco2017convolutional}.
Prior work has shown that visual foundation models can fall short on multi-view correspondence~\cite{el2024probing}, motivating methods that improve the 3D awareness of image models by learning view-consistent representations~\cite{yue2024improving,you2025multiview,zhou2025latent}.
Whereas these works learn view-consistent representations for image-based 3D perception, we examine whether view consistency in video diffusion representations is associated with output 3D consistency, and whether directly increasing it improves generation.

\section{Are View-Consistent Representations Associated with 3D-Consistent Generation?}\label{sec:analysis}

Recent studies suggest that enhancing the multi-view consistency of 2D visual representations improves their 3D awareness~\cite{yue2024improving} and performance on tasks requiring accurate 3D perception~\cite{michalkiewicz2025not,ma2024imagenet3d,yue2024improving,el2024probing}.
Building on this insight, we begin by investigating \textit{Q1: is view consistency of video diffusion features associated with output 3D consistency across camera-controlled video generation models?}
To this end, we conduct an empirical analysis on seven VDMs, evaluating the 3D consistency of their generated videos (Task I), and the view consistency of their representations (Task II).

As mentioned above, our study focuses on camera-controlled image-to-video generation, where the goal is to generate novel views of a scene given a camera trajectory and an input image, hence the content in the scene will exhibit view changes.
This setting underpins applications such as 3D reconstruction~\cite{szymanowicz2025bolt3d} and world generation~\cite{team2025hunyuanworld,wu2026video}, where the impact of 3D inconsistency is amplified.
It has therefore been a main focus in recent 3D-consistent video generation research~\cite{asim2025met3r,ren2025gen3c,zhang2025world,cao2025mvgenmaster,wu2025geometry}, supported by established 3D-consistency metrics~\cite{asim2025met3r,duan2025worldscore}.

\noindent{\bf Task I -- Video Generation.}
In camera-controlled image-to-video generation, the VDM is conditioned on an input image and a desired camera trajectory.
The goal is to generate a video where the input image is the first frame, and subsequent frames are consistent with the input camera trajectory.
We run existing VDMs on the same dataset to obtain generated videos.
To assess their 3D consistency, we use MEt3R~\cite{asim2025met3r}, which measures pixel reprojection error in the feature space of visual foundation models~\cite{caron2021emerging,fu2024featup} across generated frames, and has been shown to capture quality degradation due to 3D inconsistency.

\noindent{\bf Task II -- Geometric Correspondence.}
To evaluate the view consistency of the \emph{representations} within camera-controlled VDMs, we follow existing approaches~\cite{tang2023emergent,zhang2023tale,el2024probing} and use keypoint-free correspondence estimation~\cite{sun2021loftr}.
Rather than relying on a separate, potentially brittle keypoint detector, this task uses dense pixel-level features extracted from a VDM to establish cross-view geometric correspondences through feature-space nearest-neighbor search.
View-consistent representations encode the same 3D point more consistently across viewpoints, hence leading to more accurate correspondences.
Specifically, given a video and camera trajectory, we add a small amount of noise~\cite{tang2023emergent} to the video latents and run a VDM forward pass.
We obtain per-frame dense feature maps from the $L$th diffusion layer, which are bilinearly upsampled spatially (and temporally if the model's VAE has a temporal compression factor).
Results are reported as pixel-space correspondence recall: the percentage of detected correspondences whose Euclidean error falls below a fixed threshold.

\noindent{\bf Datasets.}
We use RealEstate10K (RE10K)~\cite{zhou2018stereo} and DL3DV~\cite{ling2024dl3dv} for both tasks, controlling for the impact of data distribution.
For geometric correspondence, we pseudo-label test videos using the feed-forward geometric estimation model VGGT~\cite{wang2025vggt}.
These labels are imperfect but sufficiently accurate for the analysis. Complete sampling and threshold details are provided in the Appendix.

\begin{figure}[t]
    \centering
    \includegraphics[width=\linewidth]{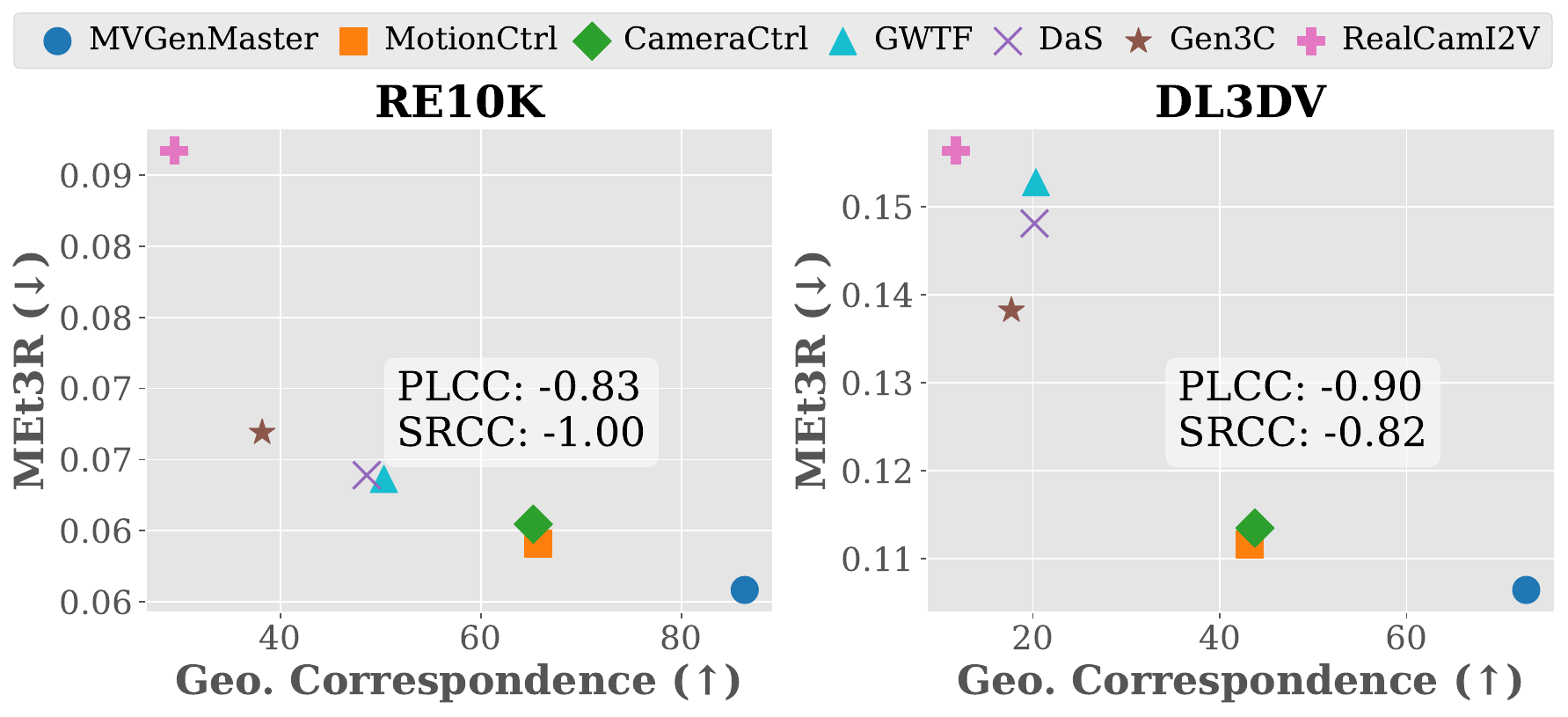}
    \vspace{-19pt}
    \caption{\textbf{Internal view consistency is strongly associated with output 3D consistency.} MEt3R and geometric-correspondence results are shown for seven camera-controlled I2V models spanning U-Net and DiT architectures on RE10K~\cite{zhou2018stereo} and DL3DV~\cite{ling2024dl3dv}.}
    \label{fig:analysis}
    \vspace{-15pt}
\end{figure}

\noindent{\bf Models.}
We evaluate CameraCtrl~\cite{he2025cameractrl}, the I2V version of MotionCtrl~\cite{wang2024motionctrl}, GEN3C~\cite{ren2025gen3c}, GWTF~\cite{burgert2025go}, DaS~\cite{gu2025diffusion}, MVGenMaster~\cite{cao2025mvgenmaster}, and RealCamI2V~\cite{li2025realcam}, covering contemporary DiT-based generators and diverse model scales.
We follow the official inference protocol of each model.
Full details are provided in the Appendix.

\noindent{\bf Results.}
In \cref{fig:analysis} we can observe that the 3D consistency of generated videos (MEt3R) strongly correlates with the view consistency of VDM representations.
This is also evident in terms of both Pearson's linear correlation (PLCC) and Spearman's rank correlation coefficients (SRCC).
That is, models whose internal representations are more view-consistent (indicating stronger ``3D awareness'') tend to produce more 3D-consistent videos.
The evaluated models span distinct conditioning mechanisms: GEN3C, GWTF, MVGenMaster, DaS, and RealCamI2V use external geometric predictors or explicitly reprojected geometry, whereas CameraCtrl and MotionCtrl rely on camera-pose conditioning without external geometric predictions.
They also span U-Net-based generators, including CameraCtrl and MotionCtrl, and DiT-based systems such as DaS and RealCamI2V.
This heterogeneity broadens the empirical scope of the analysis. However, because architecture, scale, training data, and conditioning vary jointly across models, the observed relationship cannot by itself establish that view consistency causes better output 3D consistency.
We therefore complement the cross-model analysis with matched within-model interventions that directly alter view consistency in diffusion representations while holding the base model fixed.

\noindent\textbf{Answer to Q1.}
Under the studied camera-controlled I2V setting, internal view consistency is a strong indicator of output 3D consistency across different architectures, conditioning mechanisms, and model scales.

\section{Do View-Consistent Diffusion Representations Improve Generation?}\label{sec:intervention}

Our investigation into Q1 in \cref{sec:analysis} uncovered a strong correlation between 3D consistency in camera-controlled I2V generation and view-consistent representations.
However, it does not establish that enhancing internal view consistency will improve generated outputs.
To further investigate the within-model impact of view-consistent representations, here we perform a controlled intervention on individual models, where we employ an additional loss during VDM training to encourage the learning of view-consistency in the internal representations of the same 3D points.
We refer to this loss as \name--\underline{\textbf{vi}}ew-\underline{\textbf{co}}nsistent \underline{\textbf{d}}iffusion \underline{\textbf{r}}epresentations.
For each base model, we compare training solely with their original diffusion objective against training additionally with \name.

\subsection{\name Supervision}\label{sec:method}

\begin{figure}[t]
    \centering
    \includegraphics[width=\linewidth]{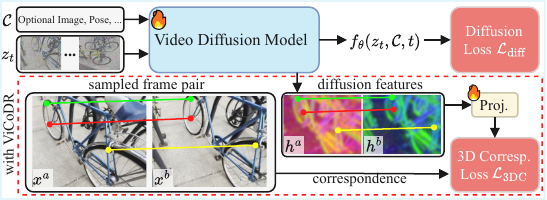}
    \vspace{-15pt}
    \caption{\textbf{Overview of \name.} During video diffusion training, \name additionally supervises internal diffusion representations $(h^a,h^b)$ extracted from frame pairs $(x^a,x^b)$ with a 3D correspondence loss (\cref{eqn:L_3dc}), so as to learn view-consistent representations (here PCA feature maps are visualized).}
    \label{fig:method}
    \vspace{-10pt}
\end{figure}

\noindent{\bf Preliminaries: Video diffusion.}
A VDM $f_\theta$, parameterized by $\theta$, aims to learn the distribution $p(x^{1:N} | \mathcal{C})$ of $N$ video frames $x^{1:N}$ with some conditioning $\mathcal{C}$, typically in a compact latent space~\cite{rombach2022high} of a pre-trained VAE $\{\mathcal{E}, \mathcal{D}\}$, with the objective:
\begin{equation}
    \mathcal{L}_\text{diff} = \mathbb{E}_{z_0, \mathcal{C}, \epsilon, t} [\norm{y - f_\theta(z_t, \mathcal{C}, t)}_2^2], \label{eqn:L_diff}
\end{equation}
where $z_0=\mathcal{E}(x^{1:N})$, and $z_t$ are the noisy video latents corrupted by random Gaussian noise $\epsilon\sim\mathcal{N}(0,I)$ at diffusion timestep $t$ following a schedule~\cite{ho2020denoising,karras2022elucidating}.
The prediction target $y$ can be computed from $\epsilon$ and/or $z_0$ using different parameterizations~\cite{ho2020denoising,song2021score,salimans2022progressive,lipman2023flow}, \eg $y=\epsilon$ in a noise-prediction setting.
The diffusion network $f_\theta$ can be implemented as a U-Net~\cite{ronneberger2015u} equipped with temporal modules or a DiT~\cite{peebles2023scalable}.
During inference, the VDM iteratively denoises the latents $z_t$, initialized as random Gaussian noise at $t=T$, and the generated video is then obtained by decoding the clean latents $x^{1:N}=\mathcal{D}(z_0)$.

\noindent{\bf Representation supervision.}
Recent works have shown that VDMs trained using a diffusion loss (\cref{eqn:L_diff}) alone often fail to produce 3D-consistent videos, resulting in objects and rigid structures deforming over time~\cite{asim2025met3r,xie2025mvgbench,jeong2025track4gen,zhang2025world}.
A plausible explanation is that, by solely learning the distribution of 2D pixels, which are lossy projections of the 3D world, these models can struggle to explain pixel transformations through the underlying scene geometry and SE(3) transforms~\cite{mitchel2024neural,wu2025geometry}.
Learning view-consistent representations may improve this missing ``3D awareness''~\cite{el2024probing,yue2024improving,you2025multiview}.
In addition to the standard diffusion loss, we therefore enforce supervision on the intermediate representations of $f_\theta$ to promote view consistency.
Specifically, we first extract per-frame feature maps $h^{1:N}\in\mathbb{R}^{N\times H_\text{ft}\times W_\text{ft}\times C}$ from noisy video latents at the $L$th layer of $f_\theta$:
\begin{equation}
    h^{1:N} = f_\theta^L(z_t^{1:N}, \mathcal{C}, t),
\end{equation}
where $H_\text{ft},W_\text{ft}$, and $C$ denote the spatial size and channel dimensions of the features.
These features are passed to a convolutional projector network $g_\phi$ and bilinearly upsampled to obtain image-resolution features $\tilde{h}^{1:N}=\mathrm{Up}[g_\phi(h^{1:N})]\in\mathbb{R}^{N\times H\times W\times C}$.
The \name loss $\mathcal{L}_\text{3DC}$, detailed below, is calculated using these projected features, with its gradients back-propagated into $f_\theta$ up to layer $L$.
To allow the view-consistent features $\tilde{h}^{1:N}$ to influence generation more directly, we follow~\cite{jeong2025track4gen} and pass them back to the diffusion network through a zero-initialized linear layer $r_\varphi$, updating the original features after detaching the gradient:
\begin{equation}
    h^{1:N} := h^{1:N} + r_\varphi(\mathtt{StopGrad}[\tilde{h}^{1:N}]).
\end{equation}
Detaching the gradient prevents $\tilde{h}^{1:N}$ from being directly influenced by the diffusion loss, allowing it to focus on learning view consistency~\cite{jeong2025track4gen}.
During VDM training, the \name term is added to the standard diffusion objective:
\begin{equation}
    \mathcal{L}_\text{total}(\{\theta,\phi,\varphi\}) = \mathcal{L}_\text{diff} + \lambda \mathcal{L}_\text{3DC}, \label{eqn:L_total}
\end{equation}
where the parameters of the additional modules $g_\phi$ and $r_\varphi$ are trained together with the VDM, and $\lambda$ weights the \name term.
The additional modules are used only during training, so \name incurs no extra inference-time cost.

\noindent{\bf View consistency loss.}
To learn view-consistent representations, we assume a static scene and employ a ranking-based 3D correspondence loss~\cite{you2025multiview}.
Having computed $\tilde{h}^{1:N}$, we randomly sample a pair of frames $(x^a,x^b)$ and their corresponding features $(\tilde{h}^a,\tilde{h}^b)$, normalizing the latter to obtain $(\bar{h}^a,\bar{h}^b)$ for cosine-similarity calculation.
We sample $K$ pixel locations as keypoints and treat each keypoint from frame $x^a$ as a query $q$.
For each $q$, we retrieve the keypoint $p$ in frame $x^b$ with the smallest 3D Euclidean distance, discarding queries for which this distance exceeds a threshold $t_\text{pos}$.
For each valid query, we also form a negative set $\mathcal{S}_\text{n}$ from keypoints in $x^b$ whose 3D distance to $q$ exceeds $t_\text{neg}$.
The 3D correspondence loss for each query $q$ is computed as $\mathcal{L}_{\text{3DC}}^q=$
\begin{equation}
1-\frac{1}{|\mathcal{S}_\text{p}|}\sum_{i\in\mathcal{S}_\text{p}}
\frac{1+\sum_{j\in\mathcal{S}_\text{p},j\neq i}\sigma_\tau(D^q_{ij})}
{1+\sum_{j\in\mathcal{S}_\text{p},j\neq i}\sigma_\tau(D^q_{ij})+\sum_{j\in\mathcal{S}_\text{n}}\sigma_\tau(D^q_{ij})},
\label{eqn:L_3dc}
\end{equation}
where $\sigma_\tau(\cdot)$ is a sigmoid function with a constant temperature $\tau=0.01$, $\mathcal{S}_\text{p}=\{q,p\}$ is the positive set including $q$ itself, and $D^q_{ij}=\bar h_q\!\cdot\!\bar h_j-\bar h_q\!\cdot\!\bar h_i$ is the difference in feature cosine similarity of the samples to $q$.
The losses for all queries are aggregated as $\mathcal{L}_\text{3DC}=\frac{1}{|\mathcal{Q}|}\sum_{q\in\mathcal{Q}}\mathcal{L}_\text{3DC}^q$, where $\mathcal{Q}$ denotes all eligible queries.
This ranking-based loss was originally proposed for image retrieval~\cite{brown2020smooth}.
The second term in \cref{eqn:L_3dc} is smooth average precision, a differentiable approximation of average precision that measures the similarity rank of a positive sample among all positive samples in the numerator and among all samples in the denominator.
It encourages a query feature to be more similar to its corresponding 3D point than to geometrically distant points, and has been shown effective for learning correspondence~\cite{you2025multiview}.

\noindent{\bf Pseudo-3D labels and sampling.}
Common datasets for training camera-controlled video generation models, such as RE10K~\cite{zhou2018stereo} and DL3DV~\cite{ling2024dl3dv}, do not contain densely annotated 3D points.
While 3D points can be recovered using optimization-based reconstruction~\cite{schoenberger2016sfm,schoenberger2016mvs,pan2024glomap}, we use the geometric prior of a large pre-trained model for efficiency and reconstruction density.
Specifically, we employ VGGT~\cite{wang2025vggt}, a feed-forward 3D reconstruction network, to densely label pixels with their corresponding 3D coordinates.
Due to its ranking-based nature, $\mathcal{L}_\text{3DC}$ constrains \textit{relative} feature similarity rather than absolute feature or coordinate values.
This relative constraint is less restrictive than direct feature distillation~\cite{wu2025geometry} or point-map regression~\cite{huang2025jog3r}, which may reduce sensitivity to imperfections in the pseudo labels. We examine this  empirically in \cref{sec:alternatives,sec:exp_ablation}.
Due to memory constraints, the sampled keypoints cannot span all pixels.
We therefore bias keypoint sampling toward Sobel edges, which capture both object contours and local texture changes and provide more spatially distinctive correspondence targets than uniform regions.
We show empirically in \cref{sec:exp_ablation} that this improves output consistency compared with uniform random sampling.

\noindent{\bf Intervention protocol.}
We apply \name to a range of video diffusion models to test whether increasing view consistency in diffusion representations improves generation across tasks.
For each intervention, the baseline and +\name variants use the same training data, initialization, trainable modules, optimization schedule, and number of training steps. Their only objective-level difference is the additional $\mathcal{L}_\text{3DC}$ term.
In the following subsections, we describe the task-specific experimental setups and present intervention results for camera-controlled I2V, camera-controlled T2V, and multi-view generation.
Shared intervention settings, full model- and task-specific training configurations, and detailed definitions and implementation for all evaluation metrics are provided in the Appendix.

\subsection{Camera-Controlled I2V Generation}\label{sec:i2v_intervention}

\noindent{\bf Models and setup.}
We instantiate this ViCoDR intervention across architectures and conditioning mechanisms using CameraCtrl (SVD)~\cite{he2025cameractrl}, DaS~\cite{gu2025diffusion}, RealCamI2V~\cite{li2025realcam}, and Wan-CameraCtrl 1.3B~\cite{wan2025wan}.
DaS and RealCamI2V are based on CogVideoX-5B~\cite{yang2024cogvideox}, a DiT architecture, and they both incorporate explicit 3D-aware conditioning, in addition to camera poses.
As both methods already receive explicit geometric inputs, they test whether supervising view consistency in internal diffusion representations remains complementary to 3D-aware conditioning.
All models are trained on RE10K~\cite{zhou2018stereo}, and evaluated on 300 held-out RE10K videos and the 1K subset of DL3DV~\cite{ling2024dl3dv} videos that cover more challenging scenes and camera trajectories.
We assess image/video quality with PSNR and FVD~\cite{unterthiner2018towards}, camera control with rotation/translation errors (RErr/TErr) and reconstruction success~\cite{he2025cameractrl,li2025realcam}, and output 3D consistency with MEt3R~\cite{asim2025met3r} and RPE~\cite{duan2025worldscore}.

\begin{table}[t]
\centering
\caption{\textbf{\name interventions for camera-controlled I2V.}
We compare \textcolor{imageMetric}{image/video quality}, \textcolor{camMetric}{camera-control accuracy}, and \textcolor{3dMetric}{output 3D consistency} under matched training.
Each base model is followed by its matched +\name run.
DaS and RealCamI2V already use explicit 3D-aware conditioning.}
\label{tab:i2v_interventions}
\vspace{-7pt}
\resizebox{\linewidth}{!}{
\begin{tabular}{@{}lccccccc@{}}
\toprule
\textbf{Model} & \textbf{\textcolor{imageMetric}{PSNR} ($\uparrow$)} & \textbf{\textcolor{imageMetric}{FVD} ($\downarrow$)} & \textbf{\textcolor{camMetric}{RErr} ($\downarrow$)} & \textbf{\textcolor{camMetric}{TErr} ($\downarrow$)} & \textbf{\textcolor{camMetric}{Succ \%} ($\uparrow$)} & \textbf{\textcolor{3dMetric}{MEt3R} ($\downarrow$)} & \textbf{\textcolor{3dMetric}{RPE} ($\downarrow$)}\\
\midrule
\multicolumn{8}{c}{\textbf{RE10K}}\\[-2pt]
\midrule
CameraCtrl & 16.89 & \textbf{113} & 0.80 & 2.31 & 86.3 & 0.066 & 0.385\\
+ \name & \textbf{17.04} & 115 & \textbf{0.61} & \textbf{1.88} & \textbf{92.3} & \textbf{0.059} & \textbf{0.334}\\ \cmidrule(l){1-8}
DaS & 17.23 & \textbf{108} & 0.52 & 1.65 & 93.5 & 0.056 & 0.310\\
+ \name & \textbf{17.24} & 109 & \textbf{0.51} & \textbf{1.62} & \textbf{93.8} & \textbf{0.053} & \textbf{0.295}\\ \cmidrule(l){1-8}
RealCamI2V & 17.36 & 104 & 0.47 & 1.50 & 94.4 & 0.052 & 0.286\\
+ \name & \textbf{17.42} & \textbf{103} & \textbf{0.45} & \textbf{1.43} & \textbf{95.0} & \textbf{0.049} & \textbf{0.268}\\ \cmidrule(l){1-8}
Wan 1.3B & \textbf{17.86} & \textbf{97} & 0.67 & 1.96 & 90.8 & 0.061 & 0.347\\
+ \name & 17.68 & 100 & \textbf{0.42} & \textbf{1.38} & \textbf{95.8} & \textbf{0.047} & \textbf{0.254}\\
\midrule
\multicolumn{8}{c}{\textbf{DL3DV}}\\[-2pt]
\midrule
CameraCtrl & 13.32 & \textbf{113} & 2.12 & 4.62 & 63.4 & 0.120 & 0.662\\
+ \name & \textbf{13.47} & 119 & \textbf{2.02} & \textbf{4.14} & \textbf{69.1} & \textbf{0.110} & \textbf{0.586}\\ \cmidrule(l){1-8}
DaS & 16.48 & \textbf{112} & 1.08 & 2.92 & 74.7 & 0.087 & 0.452\\
+ \name & \textbf{16.51} & 112 & \textbf{1.04} & \textbf{2.81} & \textbf{75.6} & \textbf{0.083} & \textbf{0.431}\\ \cmidrule(l){1-8}
RealCamI2V & 16.60 & 110 & 1.00 & 2.75 & 76.0 & 0.084 & 0.440\\
+ \name & \textbf{16.66} & \textbf{109} & \textbf{0.95} & \textbf{2.60} & \textbf{77.2} & \textbf{0.079} & \textbf{0.415}\\ \cmidrule(l){1-8}
Wan 1.3B & \textbf{17.18} & \textbf{104} & 1.17 & 3.05 & 73.9 & 0.090 & 0.472\\
+ \name & 17.12 & 108 & \textbf{0.83} & \textbf{2.28} & \textbf{79.7} & \textbf{0.072} & \textbf{0.378} \\
\bottomrule
\end{tabular}}
\vspace{-5pt}
\end{table}

\begin{table}[t]
\centering
\caption{\textbf{\name interventions for camera-controlled T2V on RE10K.}}
\label{tab:t2v_interventions}
\vspace{-7pt}
\resizebox{\linewidth}{!}{
\begin{tabular}{@{}lcccccc@{}}
\toprule
\textbf{Model} & \textbf{\textcolor{imageMetric}{FVD} ($\downarrow$)} & \textbf{\textcolor{camMetric}{RErr} ($\downarrow$)} & \textbf{\textcolor{camMetric}{TErr} ($\downarrow$)} & \textbf{\textcolor{camMetric}{Succ \%} ($\uparrow$)} & \textbf{\textcolor{3dMetric}{MEt3R} ($\downarrow$)} & \textbf{\textcolor{3dMetric}{RPE} ($\downarrow$)}\\
\midrule
MotionCtrl & 780 & \textbf{1.119} & \textbf{4.676} & 89.1 & 0.075 & 0.306\\
+ \name & \textbf{559} & 1.217 & 5.288 & \textbf{96.4} & \textbf{0.059} & \textbf{0.255}\\
\midrule
UCPE & \textbf{520} & 0.569 & 1.257 & 95.0 & 0.055 & 0.300\\
+ \name & 539 & \textbf{0.509} & \textbf{1.139} &
\textbf{95.8} & \textbf{0.049} & \textbf{0.266}\\
\bottomrule
\end{tabular}}
\vspace{-5pt}
\end{table}

\begin{table}[t]
\centering
\caption{\textbf{\name intervention for multi-view generation on OmniObject3D.}}
\label{tab:mv_intervention}
\vspace{-7pt}
\resizebox{\linewidth}{!}{
\begin{tabular}{@{}lccccc@{}}
\toprule
\textbf{Model} & \textbf{\textcolor{3dMetric}{CD} ($\downarrow$)} & \textbf{\textcolor{3dMetric}{Depth} ($\downarrow$)} & \textbf{\textcolor{imageMetric}{cPSNR} ($\uparrow$)} & \textbf{\textcolor{imageMetric}{cSSIM} ($\uparrow$)} & \textbf{\textcolor{imageMetric}{cLPIPS} ($\downarrow$)}\\
\midrule
ViFiGen & 2.98 & 11.66 & 28.91 & 0.92 & \textbf{0.04}\\
+ \name & \textbf{2.97} & \textbf{11.59} & \textbf{29.03} & \textbf{0.94} & 0.05\\
\bottomrule
\end{tabular}}
\vspace{-12pt}
\end{table}

\begin{figure*}[t]
\centering
\includegraphics[width=\linewidth]{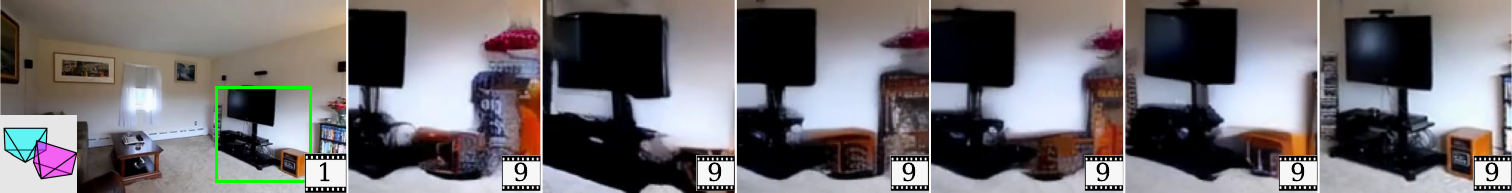}
\includegraphics[width=\linewidth]{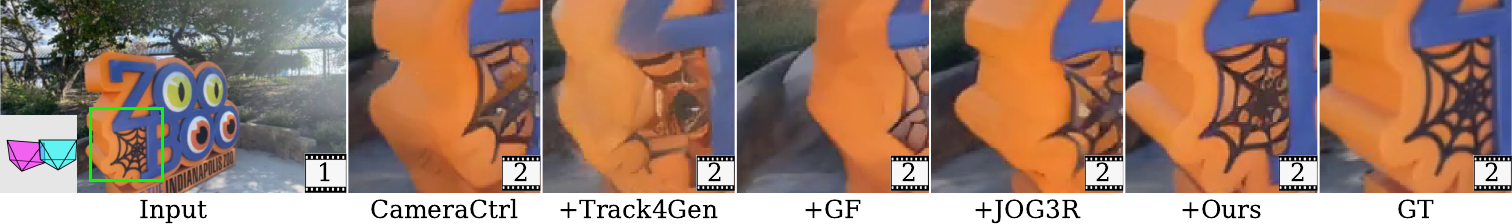}
\vspace{-18pt}
\caption{\textbf{Camera-controlled I2V comparison on CameraCtrl.}
Training with \name reduces geometric deformation while respecting the conditioning camera trajectory.
Frame indices and camera poses are shown in each example.}
\label{fig:i2v_qualitative}
\vspace{-10pt}
\end{figure*}

\noindent{\bf Results.}
As shown in \cref{tab:i2v_interventions}, on CameraCtrl, \name improves MEt3R and RPE by 11\% and 13\% on RE10K and by 8\% and 11\% on DL3DV.
The intervention also improves PSNR and all reported camera-control measurements on both datasets, although FVD increases slightly.
These quantitative improvements are supported by \cref{fig:i2v_qualitative}, where CameraCtrl trained with \name (Ours) exhibits less geometric deformation while still respecting the conditioning camera poses.
For DaS and RealCamI2V, training with \name improves both output-consistency metrics on both datasets while preserving or improving image quality and camera control.
On Wan 1.3B, \name likewise improves both output-consistency metrics and all camera-control measurements on both datasets, while slightly decreasing PSNR and increasing FVD.
These results show that increasing view consistency within diffusion models complements explicit 3D-aware input conditioning and improves the 3D consistency of videos generated by camera-controlled I2V models.

\subsection{Camera-Controlled T2V Generation}\label{sec:t2v_intervention}

\noindent{\bf Models and setup.}
We evaluate, on the same held-out RE10K as above, the AnimateDiff-based MotionCtrl~\cite{wang2024motionctrl} and the recent UCPE~\cite{zhang2026unified} using matched baseline and +\name trained on RE10K.
As these models generate scene content from text and a camera trajectory without a conditioning image, we report means over three inference seeds.
We report FVD, camera error, reconstruction success, MEt3R, and RPE.

\begin{figure}[t]
\centering
\includegraphics[width=\linewidth]{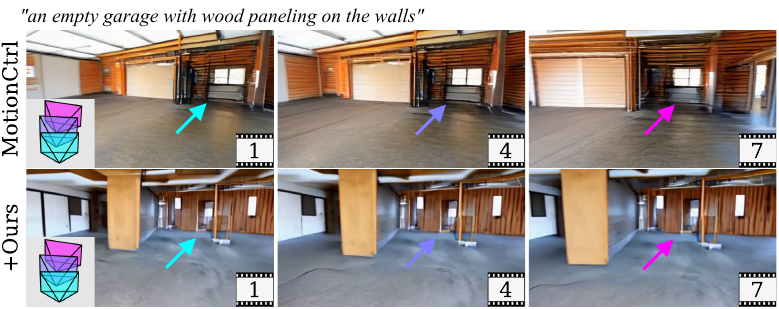}
\vspace{-15pt}
\caption{\textbf{Camera-controlled T2V comparison on MotionCtrl.}
Arrows highlight 3D-consistent and 3D-inconsistent regions.
See the Appendix for additional examples.}
\label{fig:motionctrl_qualitative}
\end{figure}

\noindent{\bf Results.}
As shown in \cref{tab:t2v_interventions}, for MotionCtrl, training with \name substantially improves FVD and both 3D-consistency metrics: MEt3R and RPE.
Reconstruction success also improves, but RErr and TErr increase slightly.
This mixed camera-control result shows that improved output consistency and more frequent successful reconstruction do not necessarily imply more precise adherence to the requested trajectory (discussed further in \cref{sec:exp_tradeoff}).
On UCPE, \name improves both output-consistency metrics and all camera-control measurements, while slightly increasing FVD.
The qualitative examples in \cref{fig:motionctrl_qualitative} further show that structures are better preserved when the model is trained with \name.

\subsection{Multi-View Generation}\label{sec:mv_intervention}

\noindent{\bf Model and setup.}
ViFiGen~\cite{xie2025mvgbench} takes an object image and generates views at fixed target cameras.
We fine-tune matched ViFiGen variants from the same SV3D backbone~\cite{voleti2024sv3d} on Objaverse~\cite{deitke2024objaverse} and evaluate them on the 6K-object OmniObject3D subset~\cite{wu2023omniobject3d}.
Following MVGBench~\cite{xie2025mvgbench}, we report Chamfer distance and rendered depth error for geometric consistency and cPSNR, cSSIM, and cLPIPS for appearance consistency.

\begin{figure}[t]
\centering
\includegraphics[width=0.9\linewidth]{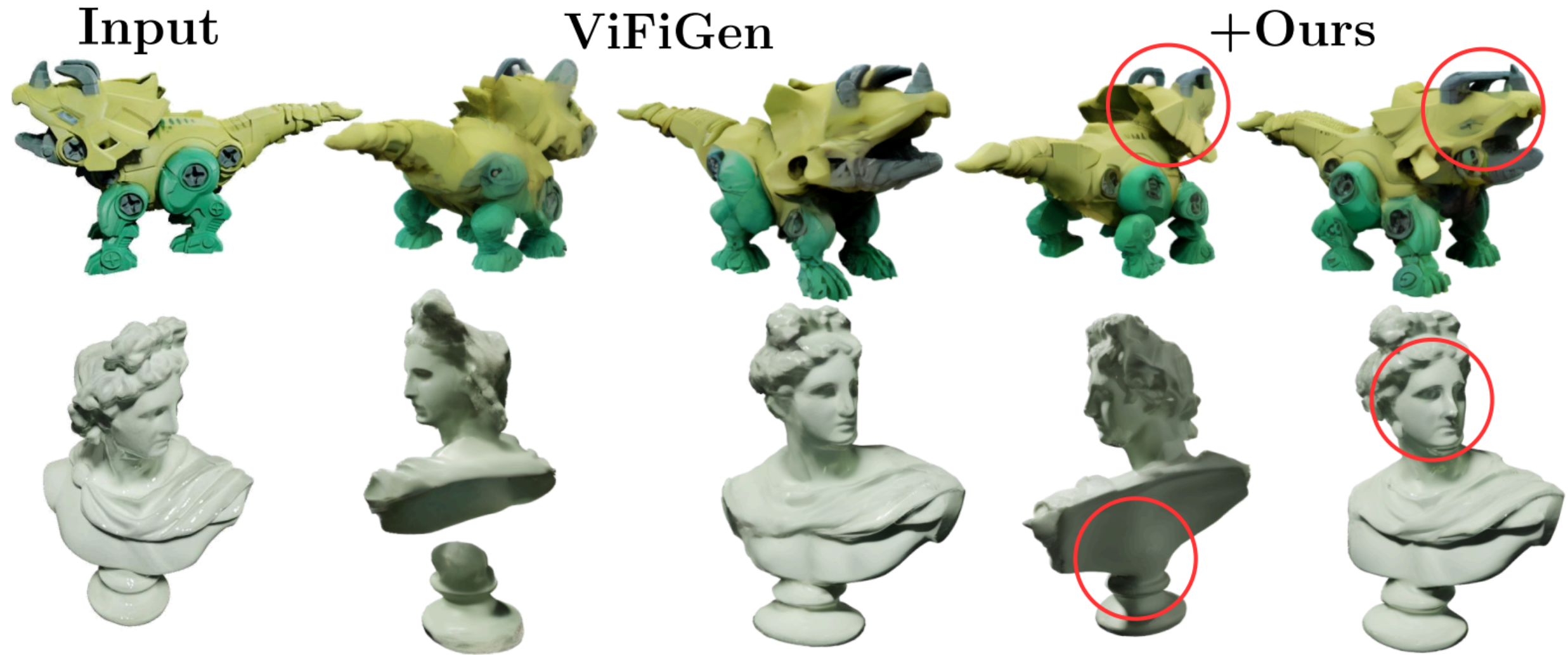}
\caption{\textbf{Multi-view generation comparison on ViFiGen.} Training with \name maintains consistent geometry.}
\label{fig:mv_qualitative}
\vspace{-10pt}
\end{figure}

\noindent{\bf Results.}
As shown in \cref{tab:mv_intervention}, training with \name lowers Chamfer distance and depth error and improves cPSNR and cSSIM, indicating more consistent geometry and appearance across viewpoints.
The slight cLPIPS degradation shows that the appearance gain is not uniform across all perceptual metrics.
Visually, \cref{fig:mv_qualitative} shows that \name produces views with stronger structural and textural coherence, including in partially occluded regions.
Ground-truth-referenced appearance metrics are in the Appendix.

\noindent\textbf{Answer to Q2.}
Across three generation tasks and multiple base models, matched training with the \name objective consistently improves the 3D-consistency measurements.

\section{Is the \name Loss an Effective and Robust Training Objective?}\label{sec:q3}

Q2 uses matched \name interventions to test whether increasing view consistency in diffusion representations improves generation across models and tasks.
Here, we evaluate the comparative effectiveness and robustness of the loss used for these interventions as an auxiliary training objective.
Using CameraCtrl as a common testbed, we compare the \name loss with existing methods for supervising diffusion representations and analyze its formulation, pseudo-label dependence, loss-weight behavior, trade-offs, and computational cost.

\subsection{Comparison with Existing Alternatives}\label{sec:alternatives}

We compare with Track4Gen~\cite{jeong2025track4gen} and JOG3R~\cite{huang2025jog3r}, which train a VDM while learning to track points and predict 3D point maps, respectively.
We also compare with Geometry Forcing (GF)~\cite{wu2025geometry}, which aligns VDM representations to VGGT features during training.
All methods use the same CameraCtrl training setup on RE10K for 50K updates.
We use the official GF implementation.
Track4Gen and JOG3R are implemented from their paper descriptions as code was unavailable.
Full optimizer, batch-size, and hardware details are provided in the Appendix.

\begin{table}[t]
\centering
\caption{\textbf{Alternative auxiliary objectives on CameraCtrl.}
Improvements from GF, JOG3R, and Track4Gen are mixed across metrics and datasets; \name is the only evaluated objective that improves both consistency metrics on both datasets.}
\label{tab:cameractrl_baseline}
\vspace{-5pt}
\resizebox{\linewidth}{!}{
\begin{tabular}{@{}clccccccc@{}}
\toprule
& \textbf{Model} & \textbf{\textcolor{imageMetric}{PSNR} ($\uparrow$)} & \textbf{\textcolor{imageMetric}{FVD} ($\downarrow$)} & \textbf{\textcolor{camMetric}{RErr} ($\downarrow$)} & \textbf{\textcolor{camMetric}{TErr} ($\downarrow$)} & \textbf{\textcolor{camMetric}{Succ \%} ($\uparrow$)} & \textbf{\textcolor{3dMetric}{MEt3R} ($\downarrow$)} & \textbf{\textcolor{3dMetric}{RPE} ($\downarrow$)}\\
\midrule
\multirow{5}{*}{\rotatebox[origin=c]{90}{RE10K}}
& CameraCtrl & {\ul 16.888} & \textbf{113} & 0.795 & 2.313 & 86.3 & {\ul 0.066} & 0.385\\
& + Track4Gen & 16.211 & 132 & 0.852 & 2.861 & 88.7 & 0.072 & 0.398\\
& + GF & 16.845 & 117 & {\ul 0.721} & {\ul 2.234} & {\ul 90.0} & 0.067 & 0.381\\
& + JOG3R & 16.103 & 120 & 0.913 & 2.702 & 86.7 & {\ul 0.066} & {\ul 0.375}\\
& + \name & \textbf{17.043} & {\ul 115} & \textbf{0.605} & \textbf{1.883} & \textbf{92.3} & \textbf{0.059} & \textbf{0.334}\\
\midrule
\multirow{5}{*}{\rotatebox[origin=c]{90}{DL3DV}}
& CameraCtrl & {\ul 13.318} & \textbf{113} & {\ul 2.119} & 4.619 & 63.4 & 0.120 & 0.662\\
& + Track4Gen & 12.965 & 227 & 2.600 & 5.610 & 54.9 & 0.137 & 0.689\\
& + GF & 13.185 & 124 & 2.425 & 5.449 & 54.6 & 0.132 & 0.735\\
& + JOG3R & 12.971 & 142 & 2.217 & {\ul 4.345} & \textbf{69.5} & {\ul 0.113} & {\ul 0.611}\\
& + \name & \textbf{13.467} & {\ul 119} & \textbf{2.023} & \textbf{4.144} & {\ul 69.1} & \textbf{0.110} & \textbf{0.586}\\
\bottomrule
\end{tabular}}
\vspace{-15pt}
\end{table}

\noindent{\bf Results.}
\cref{tab:cameractrl_baseline} summarizes the quantitative results.
Under matched training, \name is the only auxiliary objective that improves both MEt3R and RPE relative to CameraCtrl on both RE10K and DL3DV, while retaining strong image/video quality and camera-control performance.
GF improves all three camera-control measurements and RPE on RE10K but slightly worsens MEt3R. 
On DL3DV, it degrades both consistency metrics and all three camera-control measurements.
JOG3R improves RPE on both datasets and both consistency metrics on DL3DV, but its remaining quality and camera-control changes are mixed.
Track4Gen degrades both output-consistency metrics on both datasets.
These quantitative results are further supported by \cref{fig:i2v_qualitative}, where \name generates novel scene views with the least amount of geometric deformation while still respecting the conditioning camera poses. 
An additional perceptual user study is provided in the Appendix.
We hypothesize that these differences arise from how each objective uses geometric supervision.
JOG3R directly regresses point maps from diffusion features and may therefore be sensitive to imperfections in the VGGT pseudo-labels.
GF enforces feature similarity between the VDM and VGGT, which may impose an overly strict target.
Track4Gen derives point-tracking targets from 2D optical flow~\cite{teed2020raft}, which may accumulate noise when chained into longer-term tracklets.
In contrast, \name directly promotes view consistency in diffusion representations through a relative ranking objective that is less restrictive than direct feature or coordinate regression.
To confirm that \name acts on the intended representation property, a controlled CameraCtrl analysis in the Appendix shows that it increases geometric-correspondence recall while improving output 3D consistency.

\subsection{Loss Design and Pseudo-Label Dependence}\label{sec:exp_ablation}

Here we quantify the impact of the \name loss $\mathcal{L}_\text{3DC}$ and ablate design components on CameraCtrl using RE10K.

\noindent{\bf Ranking-based loss.}
To validate the ranking formulation in $\mathcal{L}_\text{3DC}$, we replace it with a simpler correspondence loss that directly maximizes the cross-view feature similarity of matched keypoints in frames $x^a$ and $x^b$.
We denote this variant as ``+\name (simple)'' in \cref{tab:ablation}.

\noindent{\bf Edge-aware sampling.}
We train ``+\name (w/o edge)'' by disabling the bias toward sampling keypoints on image edges (\cref{sec:method}).

\noindent{\bf Reliance on pseudo-label quality.}
We investigate the reliance of \name on the pseudo-correspondence labels used to compute $\mathcal{L}_\text{3DC}$.
Specifically, ``+\name (DUSt3R)'' uses DUSt3R~\cite{wang2024dust3r} instead of VGGT, a similar feed-forward 3D reconstruction model that underperforms VGGT on several benchmarks~\cite{wang2025vggt}.

\begin{table}[t]
\centering
\caption{\textbf{Analysis of \name on CameraCtrl using RE10K.}
``+\name (full)'' is the default setup using VGGT pseudo-labels.}
\vspace{-5pt}
\label{tab:ablation}
\resizebox{\linewidth}{!}{
\begin{tabular}{@{}lcccccc@{}}
\toprule
\textbf{Model} & \textbf{\textcolor{imageMetric}{FVD} ($\downarrow$)} & \textbf{\textcolor{camMetric}{RErr} ($\downarrow$)} & \textbf{\textcolor{camMetric}{TErr} ($\downarrow$)} & \textbf{\textcolor{camMetric}{Succ \%} ($\uparrow$)} & \textbf{\textcolor{3dMetric}{MEt3R} ($\downarrow$)} & \textbf{\textcolor{3dMetric}{RPE} ($\downarrow$)}\\
\midrule
CameraCtrl & \textbf{113} & 0.795 & 2.313 & 86.3 & 0.066 & 0.385\\
+ \name (simple) & 118 & 0.640 & 1.950 & 91.0 & 0.061 & 0.346\\
+ \name (w/o edge) & 118 & 0.618 & 1.966 & 92.0 & 0.061 & 0.350\\
+ \name (DUSt3R) & 117 & 0.627 & 2.012 & 90.3 & 0.060 & 0.341\\
+ \name (full) & 115 & \textbf{0.605} & \textbf{1.883} & \textbf{92.3} & \textbf{0.059} & \textbf{0.334}\\
\bottomrule
\end{tabular}}
\vspace{-10pt}
\end{table}

\noindent{\bf Results.}
As shown in \cref{tab:ablation}, the simpler correspondence loss is less effective than $\mathcal{L}_\text{3DC}$, but still improves both 3D-consistency metrics and all camera-control metrics over CameraCtrl, supporting the value of view-consistent diffusion representations.
Removing edge-aware sampling degrades both 3D-consistency metrics, indicating the importance of emphasizing object boundaries and textural details.
Qualitative results illustrating the effect of edge-aware sampling are provided in the Appendix.
Overall, the ranking formulation and edge-aware sampling provide modest complementary gains, while the small degradation with DUSt3R shows that \name benefits from stronger pseudo-labels but remains relatively robust to their source.

\subsection{Trade-offs, Cost, and Limitations}\label{sec:exp_tradeoff}

\noindent{\bf Weight of $\mathcal{L}_\text{3DC}$.}
We study the impact of the \name loss by training CameraCtrl with different values of \name loss weight, $\lambda$ (\cref{eqn:L_total}).
All models in this comparison, including CameraCtrl, are trained for the same number of iterations.

We observe that every tested nonzero value of $\lambda$ yields a large MEt3R improvement over CameraCtrl, showing that the \name objective remains effective across a broad range of loss weights rather than depending on a single tuned setting.
Increasing $\lambda$ brings small further improvement in MEt3R, while RErr and TErr increase steadily. For example, on RE10K, RErr increases from 0.605 to 0.743 as $\lambda$ increases from 0.2 to 1, with TErr following the same trend.
This reveals a trade-off between the 3D consistency of generated videos and precise camera control, although the default \name model improves both over CameraCtrl.
One possible explanation is that a weaker geometric constraint leaves the model more freedom to hallucinate content that follows the prescribed camera poses at the expense of geometric correctness.
The full loss-weight sweep and per-setting results are provided in the Appendix.
There, we also show that reducing $\lambda$ can mitigate the FVD degradation while retaining most of the MEt3R improvement.

\noindent{\bf Additional training cost.}
Obtaining VGGT correspondence labels and computing $\mathcal{L}_\text{3DC}$ increase CameraCtrl training time on two NVIDIA H200 GPUs from 42 to 51 hours (approximately 21\%), but \name incurs no additional inference cost because the auxiliary modules are discarded after training.

\noindent{\bf Limitations of the \name loss.}
The \name loss assumes a static scene because it treats matched pixels as observations of the same fixed 3D point.
Dynamic objects can violate this fixed cross-view correspondence assumption and therefore remain outside the scope of the loss evaluated here.
Although training with \name improves 3D consistency under matched training, the resulting models can still produce visual artifacts.
In particular, large viewpoint changes can expose regions for which the base generator has insufficient information, and \name does not eliminate all geometric deformation in such cases.
Extended cost details, failure cases, and edge-sampling visualizations are provided in the Appendix, along with a separate discussion of the broader limitations of our empirical study.

\noindent\textbf{Answer to Q3.}
On CameraCtrl, the \name loss compares favorably with the evaluated alternative objectives and remains effective across the tested design and pseudo-label variants.
It improves both output-consistency metrics across RE10K and DL3DV, whereas the evaluated alternative losses produce mixed results.
Its ranking formulation and edge-aware sampling are beneficial, while its loss weight provides a practical way to trade 3D consistency against camera-control accuracy and video quality.
The gains persist when using DUSt3R pseudo-labels, although their magnitude depends on pseudo-label quality; obtaining them requires additional training computation but no additional inference cost.

\section{Conclusion}

We investigated whether view consistency in diffusion representations matters for camera-controlled video generation through complementary correlation and intervention studies.
Across camera-controlled I2V models, this property is strongly associated with output 3D consistency; matched \name interventions further improve task-specific consistency across I2V, T2V, and multi-view generation.
Together, these results characterize practical trade-offs in camera control, video quality, and training cost, and establish view consistency of diffusion representations as an informative analysis signal and a useful, tunable target for more 3D-consistent camera-controlled video generation.

\section*{Acknowledgments.} 
Funding was provided by ELIAI (the Edinburgh Laboratory for Integrated Artificial Intelligence) - EPSRC (EP/W002876/1).

\clearpage
{\small
\bibliographystyle{ieeenat_fullname}
\bibliography{main}}
\clearpage
\appendix
\setcounter{table}{0}
\renewcommand{\thetable}{A\arabic{table}}
\setcounter{figure}{0}
\renewcommand{\thefigure}{A\arabic{figure}}
\noindent{\Large \bf Appendix}
\vspace{8pt}

We provide additional materials to supplement the main paper.
\cref{supp:results} includes a controlled analysis of the intended representation property, a user study, complementary quantitative and qualitative evaluations, and separate discussions of the limitations of the empirical study and the \name loss.
\cref{supp:impl} provides additional details for the correlation analysis and the training and evaluation of the video generation models.

\section{Additional results and discussion}\label{supp:results}

\subsection{Does \name Improve the Intended Property?}\label{supp:improve_corr}

We test whether the \name intervention improves the intended view consistency at the representation level.
As shown in \cref{tab:representation_intervention}, under the same correspondence protocol as the main-paper correlation analysis, CameraCtrl trained with \name exhibits substantially higher geometric-correspondence recall and lower MEt3R than a matched CameraCtrl retraining.
This verifies on the controlled CameraCtrl case that the intervention changes the internal property it is designed to target, while the task-spanning intervention results in the main paper test whether the resulting output benefit transfers.

\subsection{User study}\label{supp:user_study}
In order to further validate the perceptual improvement in 3D consistency brought by \name, we performed a user study. 
We adopted a two-alternative forced-choice (2AFC) protocol to compare CameraCtrl+\name against each of four baselines: CameraCtrl and CameraCtrl trained with JOG3R, GF, or Track4Gen.
In each comparison, participants viewed a pair of videos generated from identical inputs, one from CameraCtrl+\name and the other from the corresponding baseline.
The spatial presentation ordering of the two videos was randomized. 
The test videos were sampled from the RE10K and DL3DV test sets after filtering out videos with insufficient camera movement, \ie ones in which the baseline between the first and last frames was less than 10\% of the median scene depth, where the poses and depth are obtained with VGGT~\cite{wang2025vggt}. 
Participants were asked to choose a video to answer the question ``\emph{Which video keeps shapes and structures more consistent?}''. 
A total of 53 participants completed the survey.
Ethics approval for the study was obtained from the relevant internal Institutional Review Board.

\begin{table}[t]
\centering
\caption{\textbf{Output 3D consistency and geometric correspondence for CameraCtrl trained with and without \name.}
$\dagger$ denotes the pretrained checkpoint released by~\cite{he2025cameractrl}; the other two models use matched training.}
\label{tab:representation_intervention}
\resizebox{\linewidth}{!}{
\begin{tabular}{@{}lcccc@{}}
\toprule
& \multicolumn{2}{c}{RE10K} & \multicolumn{2}{c}{DL3DV}\\
\cmidrule(lr){2-3}\cmidrule(lr){4-5}
\textbf{Model} & \textbf{\textcolor{3dMetric}{MEt3R} ($\downarrow$)} & \textbf{\textcolor{3dMetric}{Geo. Corr.} ($\uparrow$)} & \textbf{\textcolor{3dMetric}{MEt3R} ($\downarrow$)} & \textbf{\textcolor{3dMetric}{Geo. Corr.} ($\uparrow$)}\\
\midrule
CameraCtrl$^\dagger$ & 0.060 & 65.2 & 0.113 & 43.8\\
CameraCtrl & 0.066 & 57.7 & 0.120 & 36.7\\
CameraCtrl + \name & \textbf{0.059} & \textbf{91.0} & \textbf{0.110} & \textbf{75.0}\\
\bottomrule
\end{tabular}}
\end{table}

\begin{figure}[t]
    \centering
    \includegraphics[width=\linewidth]{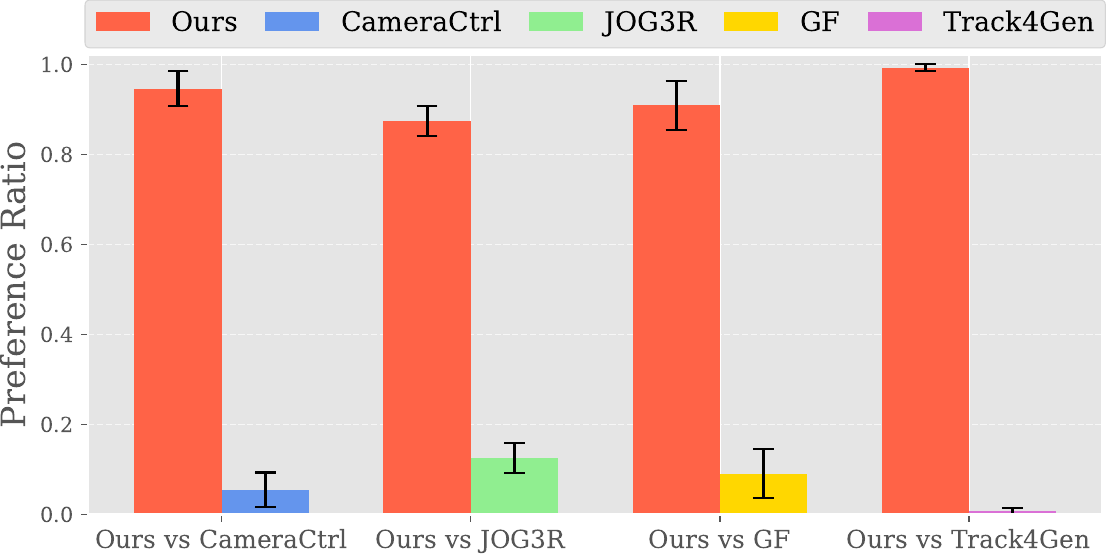}
    \vspace{-5pt} 
    \caption{\textbf{User study results.}
    Preference ratios are shown for four pairwise comparisons.
    CameraCtrl+\name is preferred substantially more often than each baseline for maintaining shape and structural consistency.}
    \label{fig:user_study}
\end{figure}

The user study results are summarized in \cref{fig:user_study}, where we report the mean and standard deviation of preference ratios for each method across sequences. 
CameraCtrl+\name is preferred substantially more often than every baseline for maintaining shape and structural consistency.
The user-study results align with the quantitative metrics and provide additional perceptual evidence for the \name intervention.

\begin{table}[t]
\centering
\caption{\textbf{Full results of the \name intervention on camera-controlled T2V generation.}
Mean results over three seeds are reported for both models; standard deviations are additionally reported for MotionCtrl.}
\label{tab:motionctrl_full}
\resizebox{\linewidth}{!}{
\begin{tabular}{@{}lcccccc@{}}
    \toprule
    \textbf{Model} & \textbf{\textcolor{imageMetric}{FVD} ($\downarrow$)} & \textbf{\textcolor{camMetric}{RErr} ($\downarrow$)} & \textbf{\textcolor{camMetric}{TErr} ($\downarrow$)} & \textbf{\textcolor{camMetric}{Succ \%} ($\uparrow$)} & \textbf{\textcolor{3dMetric}{MEt3R} ($\downarrow$)} & \textbf{\textcolor{3dMetric}{RPE} ($\downarrow$)} \\
    \midrule
    MotionCtrl       & 780 (18) & \textbf{1.119 (0.057)} & \textbf{4.676 (0.238)} & 89.1 (1.7) & 0.075 (0.001) & 0.306 (0.008) \\
    + ViCoDR   & \textbf{559 (14)}	& 1.217 (0.024) & 5.288 (0.049) & \textbf{96.4 (1.1)} & \textbf{0.059 (0.001)} & \textbf{0.255 (0.002)} \\
    \midrule
    UCPE
    & \textbf{520}
    & 0.569
    & 1.257
    & 95.0
    & 0.055
    & 0.300
    \\
    + ViCoDR
    & 539
    & \textbf{0.509}
    & \textbf{1.139}
    & \textbf{95.8}
    & \textbf{0.049}
    & \textbf{0.266}
    \\
    \bottomrule
\end{tabular}
}
\end{table}

\begin{table}[t]
    \centering
    \caption{\textbf{Additional results on multi-view generation.}
    On OmniObject3D, \name outperforms ViFiGen across all four ground-truth-referenced appearance metrics.}
    \label{tab:vifigen_supp}
    \resizebox{\linewidth}{!}{
    \begin{tabular}{@{}lcccc@{}}
        \toprule
        \textbf{Model} & \textbf{\textcolor{imageMetric}{mPSNR} ($\uparrow$)} & \textbf{\textcolor{imageMetric}{mSSIM} ($\uparrow$)} & \textbf{\textcolor{imageMetric}{mLPIPS} ($\downarrow$)} & \textbf{\textcolor{imageMetric}{FID} ($\downarrow$)} \\
        \midrule
        ViFiGen    &  19.08 & 0.77 & 0.19 & 27.87 \\
        + ViCoDR   &  \textbf{19.78} & \textbf{0.81} & \textbf{0.13} & \textbf{20.35} \\
        \bottomrule
    \end{tabular}
    }
\end{table}

\subsection{Additional quantitative results}

\noindent\textbf{CameraCtrl.} The main paper reports a compact comparison of CameraCtrl trained with \name and alternative auxiliary objectives, omitting SSIM and LPIPS for space.
\cref{tab:cameractrl_baseline_full} provides the complete nine-metric results on RE10K and DL3DV.
The additional image-level metrics reinforce the same conclusion: \name achieves the best PSNR, SSIM, and LPIPS on both datasets while remaining the only auxiliary objective that improves both MEt3R and RPE on both datasets.

\begin{table*}[t]
\centering
\caption{\textbf{Full comparison of alternative auxiliary objectives on CameraCtrl.}
This table complements the compact main-paper comparison by additionally reporting SSIM and LPIPS.}
\label{tab:cameractrl_baseline_full}
\resizebox{\linewidth}{!}{
\begin{tabular}{@{}lccccccccc@{}}
\toprule
& \multicolumn{9}{c}{RE10K}\\
\cmidrule(l){2-10}
\textbf{Model} & \textbf{\textcolor{imageMetric}{PSNR} ($\uparrow$)} & \textbf{\textcolor{imageMetric}{SSIM} ($\uparrow$)} & \textbf{\textcolor{imageMetric}{LPIPS} ($\downarrow$)} & \textbf{\textcolor{imageMetric}{FVD} ($\downarrow$)} & \textbf{\textcolor{camMetric}{RErr} ($\downarrow$)} & \textbf{\textcolor{camMetric}{TErr} ($\downarrow$)} & \textbf{\textcolor{camMetric}{Succ \%} ($\uparrow$)} & \textbf{\textcolor{3dMetric}{MEt3R} ($\downarrow$)} & \textbf{\textcolor{3dMetric}{RPE} ($\downarrow$)}\\
\midrule
CameraCtrl & {\ul 16.888} & {\ul 0.695} & {\ul 0.273} & \textbf{113} & 0.795 & 2.313 & 86.3 & {\ul 0.066} & 0.385\\
+ Track4Gen & 16.211 & 0.666 & 0.300 & 132 & 0.852 & 2.861 & 88.7 & 0.072 & 0.398\\
+ GF & 16.845 & 0.693 & 0.278 & 117 & {\ul 0.721} & {\ul 2.234} & {\ul 90.0} & 0.067 & 0.381\\
+ JOG3R & 16.103 & 0.670 & 0.296 & 120 & 0.913 & 2.702 & 86.7 & {\ul 0.066} & {\ul 0.375}\\
+ \name & \textbf{17.043} & \textbf{0.699} & \textbf{0.266} & {\ul 115} & \textbf{0.605} & \textbf{1.883} & \textbf{92.3} & \textbf{0.059} & \textbf{0.334}\\
\bottomrule
\end{tabular}}
\par\smallskip
\resizebox{\linewidth}{!}{
\begin{tabular}{@{}lccccccccc@{}}
\toprule
& \multicolumn{9}{c}{DL3DV}\\
\cmidrule(l){2-10}
\textbf{Model} & \textbf{\textcolor{imageMetric}{PSNR} ($\uparrow$)} & \textbf{\textcolor{imageMetric}{SSIM} ($\uparrow$)} & \textbf{\textcolor{imageMetric}{LPIPS} ($\downarrow$)} & \textbf{\textcolor{imageMetric}{FVD} ($\downarrow$)} & \textbf{\textcolor{camMetric}{RErr} ($\downarrow$)} & \textbf{\textcolor{camMetric}{TErr} ($\downarrow$)} & \textbf{\textcolor{camMetric}{Succ \%} ($\uparrow$)} & \textbf{\textcolor{3dMetric}{MEt3R} ($\downarrow$)} & \textbf{\textcolor{3dMetric}{RPE} ($\downarrow$)}\\
\midrule
CameraCtrl & {\ul 13.318} & {\ul 0.493} & {\ul 0.433} & \textbf{113} & {\ul 2.119} & 4.619 & 63.4 & 0.120 & 0.662\\
+ Track4Gen & 12.965 & 0.478 & 0.479 & 227 & 2.600 & 5.610 & 54.9 & 0.137 & 0.689\\
+ GF & 13.185 & 0.491 & 0.447 & 124 & 2.425 & 5.449 & 54.6 & 0.132 & 0.735\\
+ JOG3R & 12.971 & 0.479 & 0.450 & 142 & 2.217 & {\ul 4.345} & \textbf{69.5} & {\ul 0.113} & {\ul 0.611}\\
+ \name & \textbf{13.467} & \textbf{0.499} & \textbf{0.431} & {\ul 119} & \textbf{2.023} & \textbf{4.144} & {\ul 69.1} & \textbf{0.110} & \textbf{0.586}\\
\bottomrule
\end{tabular}}
\end{table*}

\noindent\textbf{Camera-controlled T2V.}
The main paper reports the mean results of the MotionCtrl and UCPE interventions over three runs with different seeds.
For completeness, \cref{tab:motionctrl_full} additionally reports the corresponding standard deviations for MotionCtrl.

\noindent\textbf{ViFiGen.} In addition to the MVGBench metrics reported in the main paper, we also compute conventional image-level reconstruction metrics on OmniObject3D~\cite{wu2023omniobject3d} in \cref{tab:vifigen_supp}. The OmniObject3D dataset contains 6,000 scanned objects across 190 common object categories.
We render 21 target views for each test object and compare each generated image with its corresponding ground-truth view. We report mean PSNR (mPSNR), mean SSIM (mSSIM), and mean LPIPS (mLPIPS) by first averaging over views for each object and then averaging over all objects. 
For FID, we follow the standard protocol and compute the Fréchet Inception Distance (FID)~\cite{heusel2017gans} between the distribution of generated views and the corresponding ground-truth views. 
Note that this evaluation differs from the 3DGS-based self-consistency metrics of MVGBench~\cite{xie2025mvgbench}, and is closer in spirit to the GT-based multi-view evaluation used in SV3D~\cite{voleti2024sv3d}.
For fairness, the ViFiGen baseline and ViFiGen+\name use the same data and training setup, as detailed in \cref{supp:impl_vicodr}.
As shown in \cref{tab:vifigen_supp}, ViFiGen+\name outperforms ViFiGen across all four ground-truth-referenced metrics, complementing the self-consistency results in the main paper.

\subsection{Additional qualitative results} \label{supp:qualitative}
\begin{figure}[t]
    \centering
    \includegraphics[width=\linewidth]{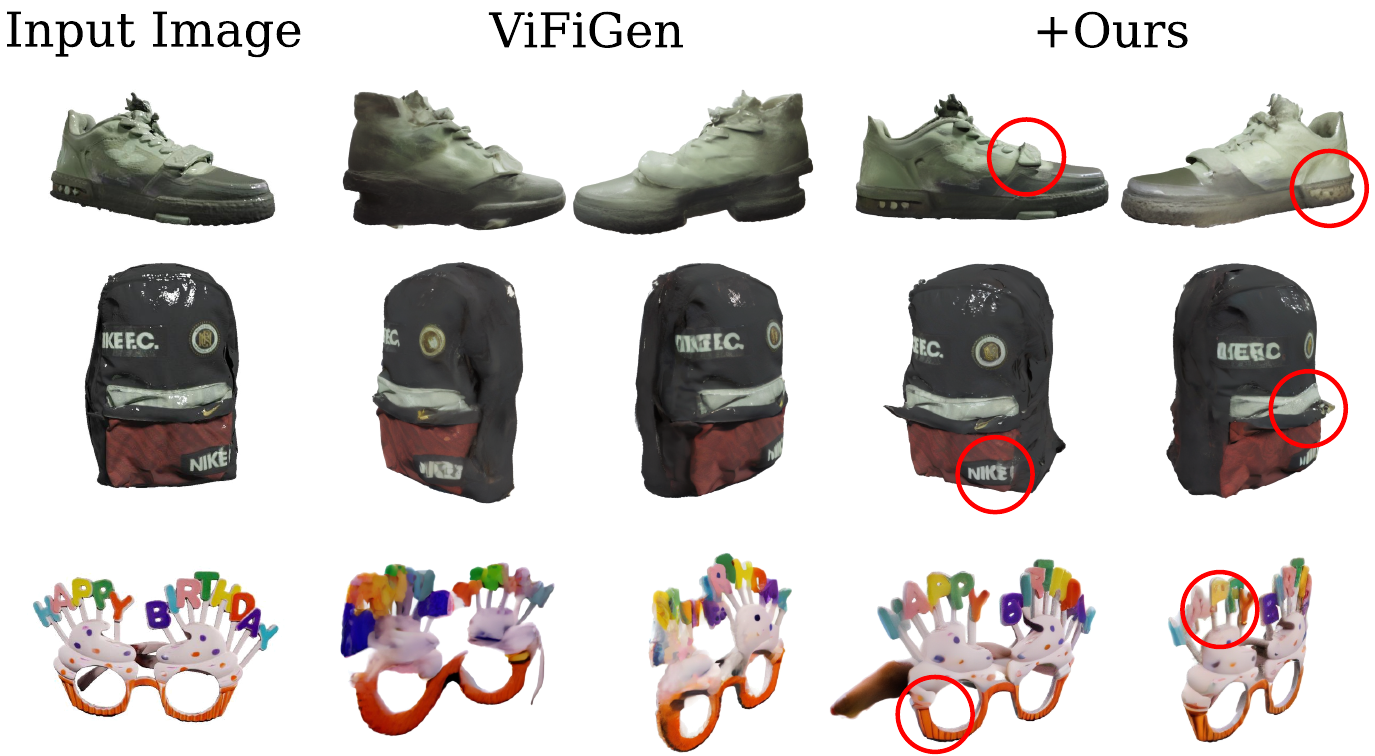}
    \caption{\textbf{Additional results on multi-view generation.} The red circles highlight regions of improved 3D consistency.}
    \label{fig:visual_vifigen_supp}
\end{figure}

\begin{figure*}[t]
    \centering
    \includegraphics[width=0.93\linewidth]{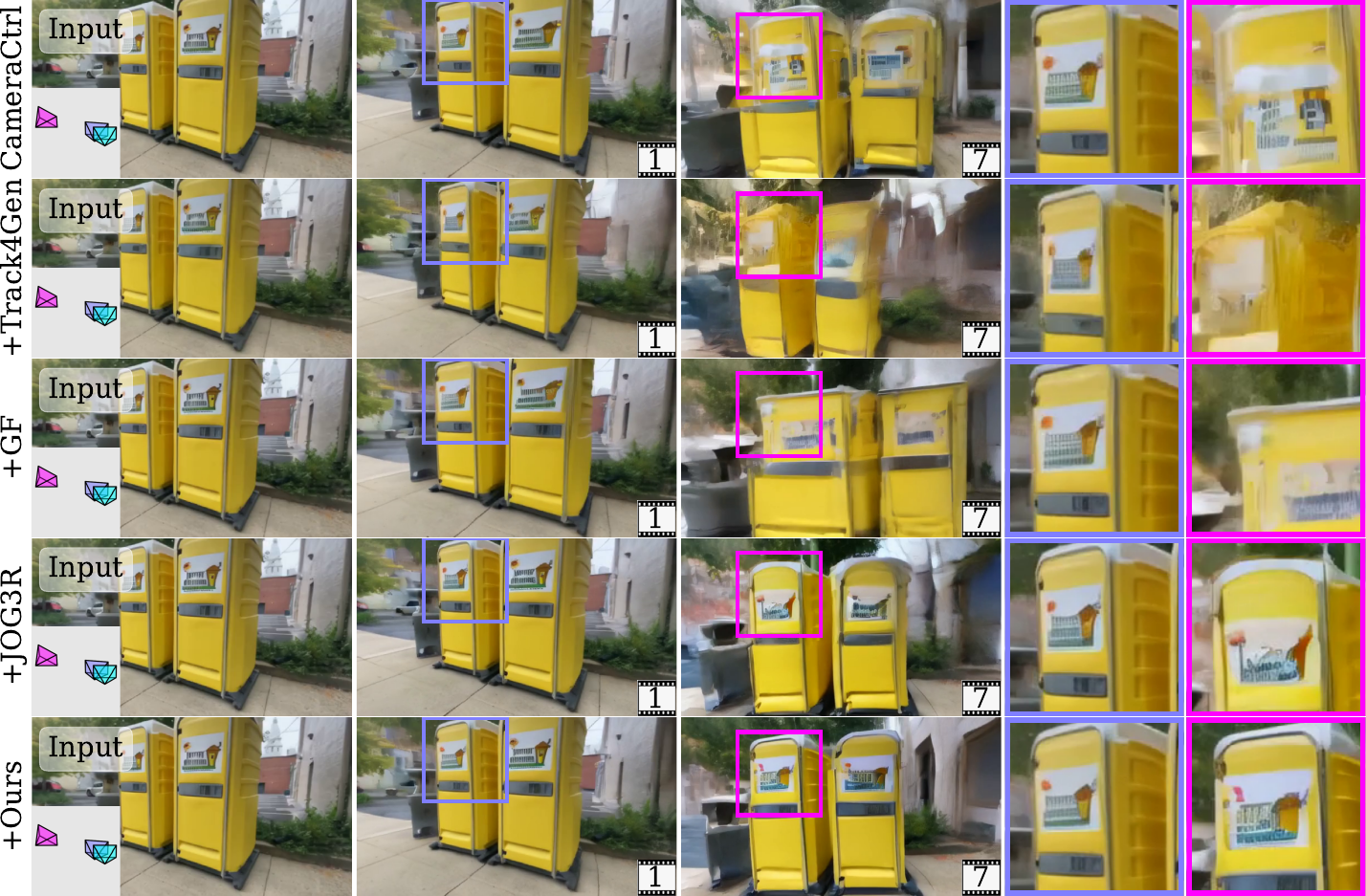} \\
    \includegraphics[width=0.93\linewidth]{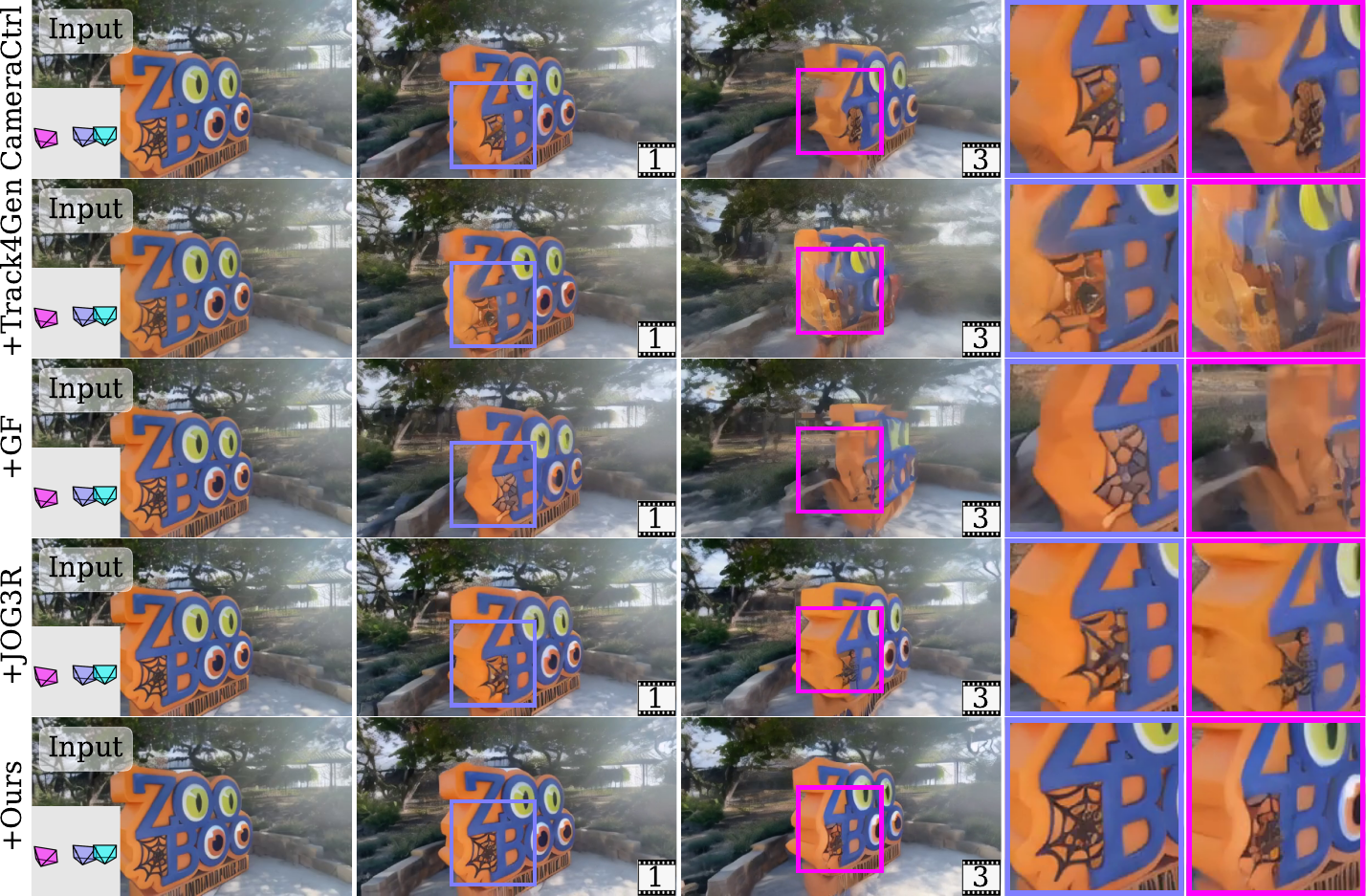}
    \caption{\textbf{Additional qualitative results on camera controlled image-to-video generation.} Given the \textcolor{lightblue}{\textbf{input frame}},  \name better maintains 3D consistency across \textcolor{darkerblue}{\textbf{view one}} and \textcolor{brightpurple}{\textbf{view two}} compared to the baselines. 
    }
    \label{fig:visual_cameractrl_1}
\end{figure*}

\begin{figure*}[t]
    \centering
    \includegraphics[width=0.93\linewidth]{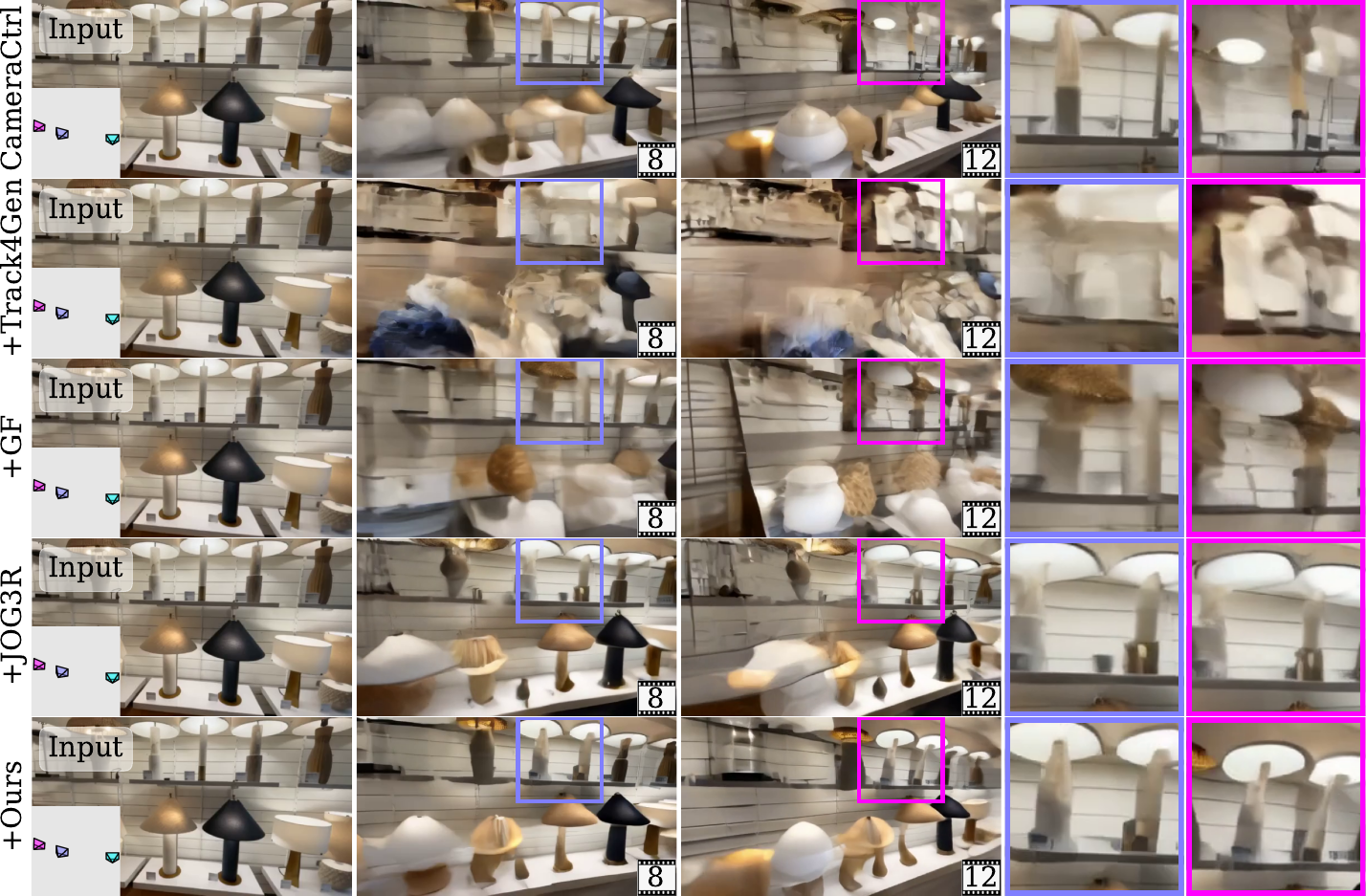} \\
    \includegraphics[width=0.93\linewidth]{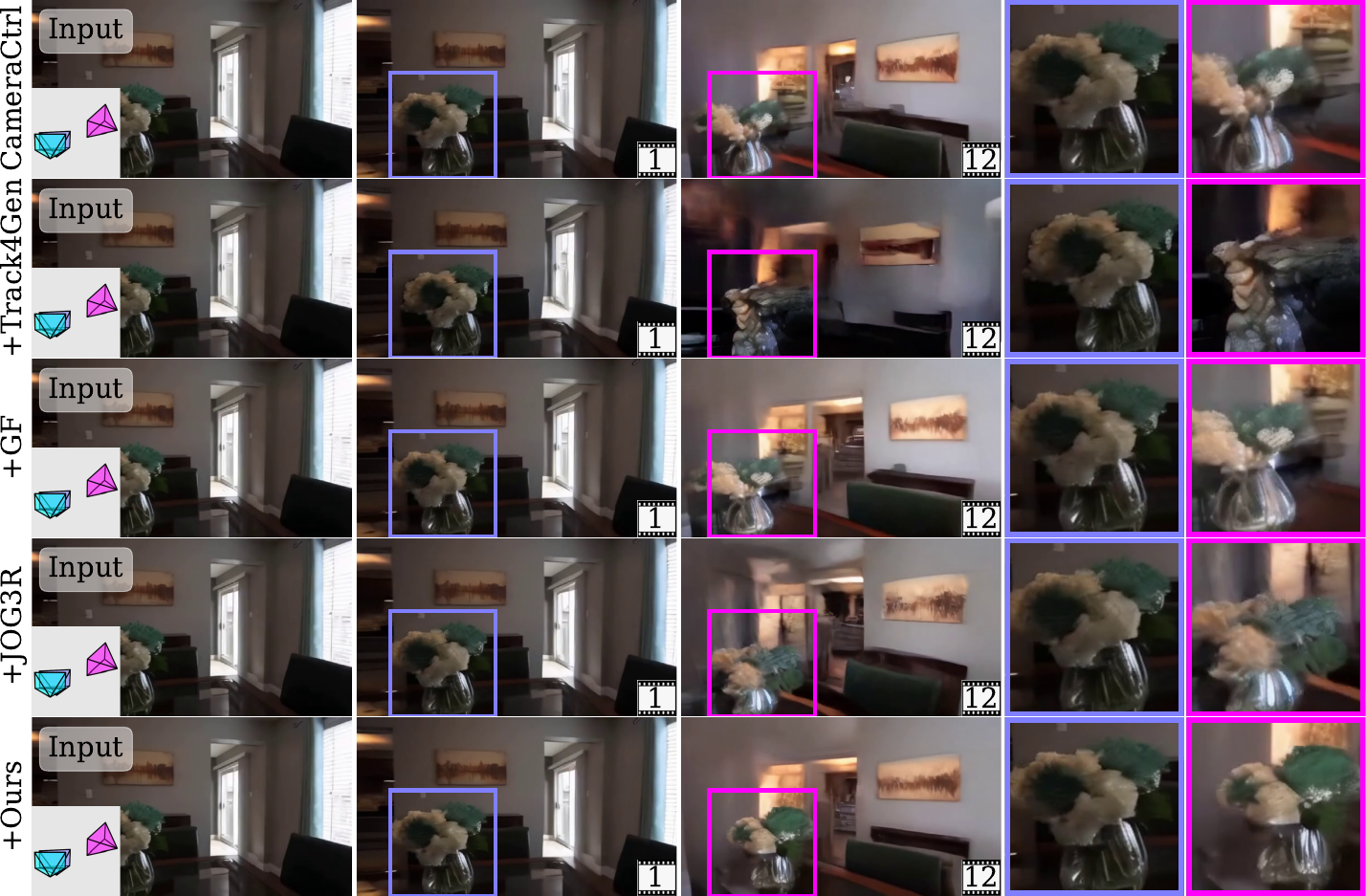}
    \caption{\textbf{Additional qualitative results on camera controlled image-to-video generation.} Given the \textcolor{lightblue}{\textbf{input frame}},  \name better maintains 3D consistency across \textcolor{darkerblue}{\textbf{view one}} and \textcolor{brightpurple}{\textbf{view two}} compared to the baselines. 
    }
    \label{fig:visual_cameractrl_2}
\end{figure*}

\begin{figure*}[t]
    \centering
    \includegraphics[width=0.8\linewidth]{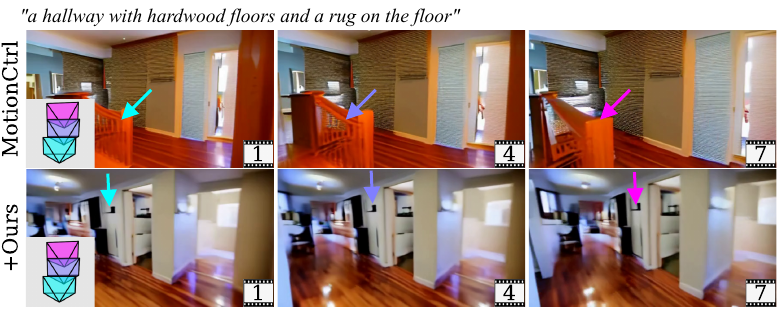}
    \includegraphics[width=0.8\linewidth]{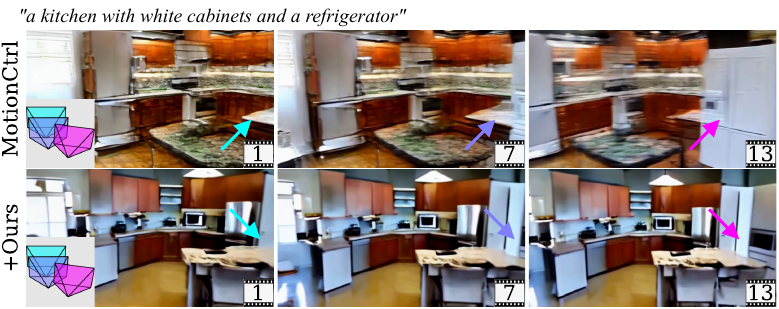}
    \includegraphics[width=0.8\linewidth]{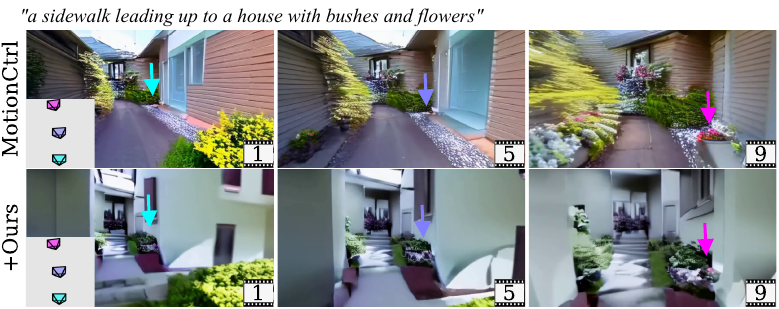}
    \caption{\textbf{Qualitative results of applying \name to text-to-video generation.} 
    The input text prompt is shown above the images. 
    Arrows highlight 3D consistent/inconsistent regions.}
    \label{fig:visual_motionctrl_supp}
\end{figure*}

Here we provide additional qualitative results of applying \name to the three base models: CameraCtrl (\cref{fig:visual_cameractrl_1}, \cref{fig:visual_cameractrl_2}), MotionCtrl (\cref{fig:visual_motionctrl_supp}), and ViFiGen (\cref{fig:visual_vifigen_supp}).
It can be observed from these results that \name better preserves object shape and texture across multiple views compared to the existing alternatives.

\subsection{Additional details on DiT-based 3D-aware VDMs}
This section provides additional context for the camera-controlled I2V interventions on models that (i) use larger DiT~\cite{peebles2023scalable} backbones and (ii) incorporate explicit 3D-aware conditioning in addition to camera poses.
These models are DaS~\cite{gu2025diffusion} and RealCamI2V~\cite{li2025realcam}, both of which use the DiT-based CogVideoX-5B I2V backbone~\cite{yang2024cogvideox}.

\noindent\textbf{DaS.} Given a conditioning input image and a sequence of camera poses, DaS firstly employs a monocular depth estimator to project the input image to 3D points. The 3D points are colored with their 3D coordinates then projected onto the input camera trajectory, constructing a ``3D tracking video'', which provides a more 3D-aware control signal to the generation process, in addition to the camera poses.

\noindent\textbf{RealCamI2V.} RealCamI2V also estimates depth from the input image, and projects it into a point cloud. Then, the point cloud is projected onto the input camera trajectory to obtain a ``preview'' of the desired video following the camera path. This is similar to GEN3C~\cite{ren2025gen3c} where a ``3D cache'' (\ie a point cloud) is rendered at the desired camera poses. Once the ``preview'' is obtained, the VDM is then conditioned on the input image, camera trajectory and the ``preview'' video to generate the final video. 

Both methods utilize explicit 3D information as additional input to the VDM in order to improve 3D consistency.
The main-paper I2V intervention table reports their matched baseline and +\name results; the descriptions above clarify the explicit 3D-aware conditioning to which the \name objective is complementary.

\begin{figure*}[t]
    \centering
    \includegraphics[width=\linewidth]{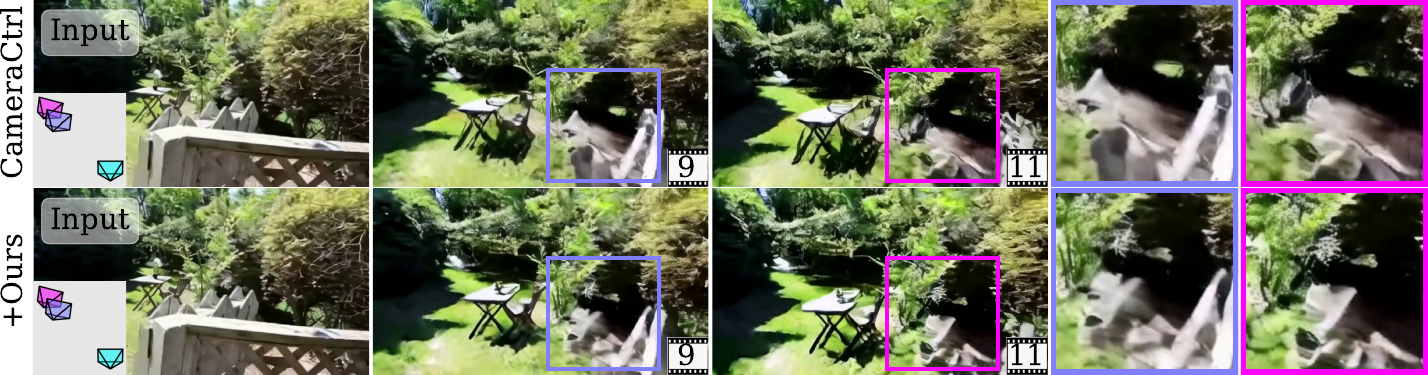} \\
    \caption{\textbf{Failure case 1: very large view changes.} Given the \textcolor{lightblue}{\textbf{input frame}}, \name fails to improve the 3D consistency of CameraCtrl in the highlighted region across \textcolor{darkerblue}{\textbf{view one}} and \textcolor{brightpurple}{\textbf{view two}}, which is exposed to large view changes. \name produces clearer edges.
    }
    \label{fig:visual_failure_1}
\end{figure*}

\begin{figure*}[t]
    \centering
    \includegraphics[width=\linewidth]{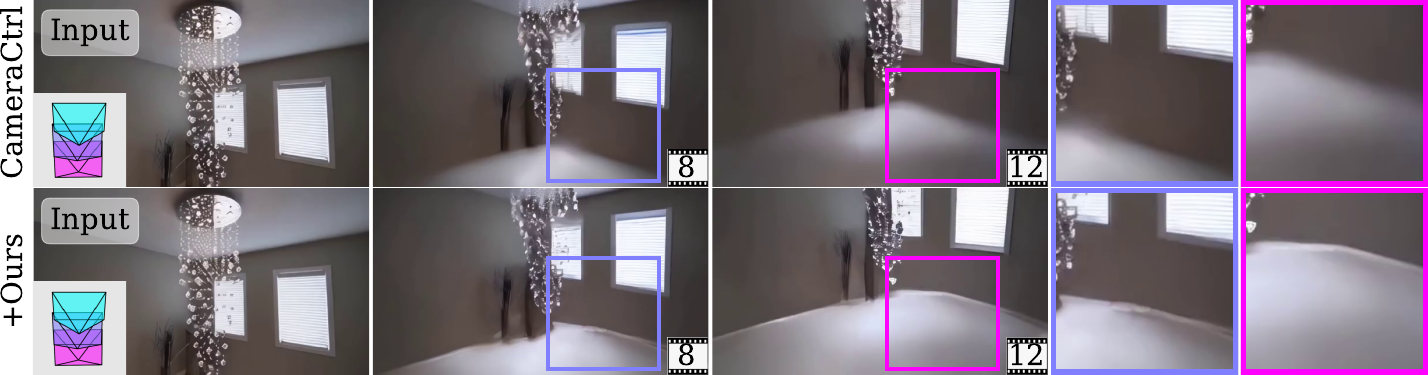} \\
    \caption{\textbf{Failure case 2: clearer but distorted edges.} Given the \textcolor{lightblue}{\textbf{input frame}}, compared to CameraCtrl, \name produces clearer edges across \textcolor{darkerblue}{\textbf{view one}} and \textcolor{brightpurple}{\textbf{view two}}. However, this can appear less natural compared to the smeared edges in the CameraCtrl output.
    }
    \label{fig:visual_failure_2}
\vspace{-10pt}
\end{figure*}

\subsection{Limitations of the empirical study}\label{supp:study_limitations}

Our empirical conclusions have several scope limitations.
First, Q1 measures cross-model association and does not isolate causality because architecture, scale, training data, and conditioning vary jointly, although the matched Q2 interventions provide complementary within-model evidence.
Second, Q2 covers multiple models, architectures, and generation tasks but is not an exhaustive evaluation of camera-controlled video generators.
Third, the alternative-objective comparison, ablations, trade-offs, and cost analysis in Q3 use CameraCtrl as a controlled testbed, so their quantitative behavior may not transfer unchanged to other models.
Finally, our experiments predominantly cover static scenes, and we leave extended study on dynamic scenes using 4D reconstruction methods (\eg \cite{wang2026vggt}) as future work.

\subsection{Limitations and failure cases of \name}\label{supp:limitations}
In addition to the limitations of the \name loss summarized in the main paper, here we discuss its performance failure modes and additional training cost in greater detail.

\noindent\textbf{Failure case 1.} It is observed that, under the same training iterations, although the base models trained with \name show improved 3D consistency compared to when they are trained without \name (supported by both quantitative and qualitative evaluation), they can still produce visual artifacts.
\cref{fig:visual_failure_1} shows an example of this, where \name fails to improve the 3D consistency of CameraCtrl in the fence area across large view changes. 
As discussed in the main paper, this can be partially due to the inherent limitations of the base model.
We hypothesize that some of these failures reflect limitations of CameraCtrl's Stable Video Diffusion backbone, whose mostly convolutional U-Net architecture may be weaker at modeling long-range spatio-temporal dependencies than larger transformer-based architectures~\cite{peebles2023scalable}.
The main-paper interventions on DiT-based models broaden the evidence beyond this backbone, but they do not imply that \name eliminates base-model failures under large viewpoint changes.

\noindent\textbf{Failure case 2.} As can be seen from the qualitative examples (\cref{supp:qualitative}), \name generally preserves edges more clearly compared to the baseline CameraCtrl (likely due to our edge-aware sampling strategy during training, see \cref{supp:ablation_edge}). 
While this is often perceived as improved 3D consistency, in some cases this can lead to lower perceptual quality of the video.
This is illustrated in \cref{fig:visual_failure_2}, where CameraCtrl+\name generates sharper and clearer edges at the corner of the walls than CameraCtrl.
However, these lines appear to violate linear perspective and are not perfectly straight.
In contrast, the slightly blurry shadows generated by the baseline CameraCtrl look more natural and can be perceived as being of higher quality.
This indicates failed attempts to improve 3D consistency in video generation can even undermine the perceptual quality. 
An interesting avenue for future research is to study the effect of improved 3D consistency on perceptual quality in video generation.

\noindent\textbf{Additional training cost.} 
Training a video generation model with \name requires additional computation beyond default diffusion training, mainly from using VGGT to obtain correspondence labels and computing the $\mathcal{L}_\text{3DC}$ objective defined in the main paper.
On our hardware setup (detailed in \cref{supp:impl_vicodr}), this results in a 21\% increase in training time for CameraCtrl, extending 42-hour training time of CameraCtrl to 51 hours, enabling feasible scaling with modest compute.

\noindent\textbf{Discussion on CameraCtrl FVD results.}
In the main-paper CameraCtrl comparison, CameraCtrl+\name has slightly worse FVD than CameraCtrl.
As discussed further in \cref{supp:ablation_edge}, a possible reason is that the 3D correspondence loss constrains generative flexibility in unseen regions.
We trained and tested CameraCtrl+\name with $\lambda = 0.15$ on RE10K; FVD decreases to 111 (CameraCtrl: 113), while MEt3R remains improved at 0.060 (CameraCtrl: 0.066; default CameraCtrl+\name: 0.059).
This suggests that reducing $\lambda$ can mitigate FVD degradation while retaining an improvement in 3D consistency.

\subsection{Additional analysis of \name}\label{supp:ablation_edge}

\begin{figure}[t]
    \centering
    \includegraphics[width=\linewidth]{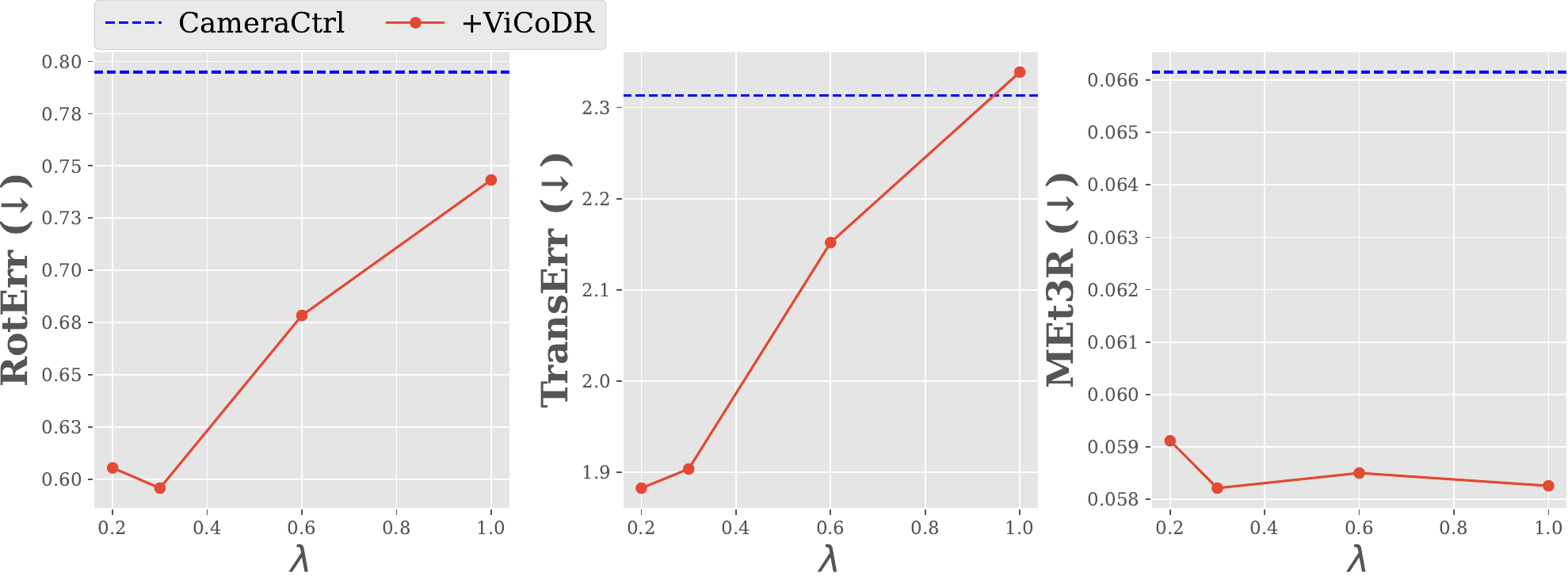} 
    \vspace{-15pt}
    \caption{\textbf{Impact of $\lambda$ on RE10K performance.}
    Increasing $\lambda$ further reduces MEt3R but increases RErr and TErr, revealing a trade-off between output 3D consistency and camera-control accuracy; the default \name setting nevertheless improves both over CameraCtrl.}
    \label{fig:ablation_lambda}
\end{figure}

\noindent\textbf{Weight of $\mathcal{L}_\text{3DC}$.} The impact of $\mathcal{L}_\text{3DC}$ on the 3D consistency and camera control accuracy of CameraCtrl is discussed in the main paper. 
Here we provide full RE10K results for models trained with different values of $\lambda$, the weight of $\mathcal{L}_\text{3DC}$ in the total training objective defined in the main paper.
These results are summarized in \cref{fig:ablation_lambda}, where we  observe that even a small value of $\lambda$ (minimum tested was 0.1) results in a large improvement in 3D consistency (MEt3R), while increasing $\lambda$ trades off camera control preciseness for small improvements in MEt3R.
We hypothesize that this trade-off may reflect improved physical correctness: with a weaker geometric constraint (smaller $\lambda$), the model may have more freedom to ``hallucinate'' content that satisfies the requested camera poses while sacrificing geometric correctness.
Despite this trade-off, the default CameraCtrl+\name setting still improves camera control over CameraCtrl, as reported in the main-paper CameraCtrl comparison and illustrated by \cref{fig:ablation_lambda}.

\begin{figure}[!t]
    \centering
    \includegraphics[width=\linewidth]{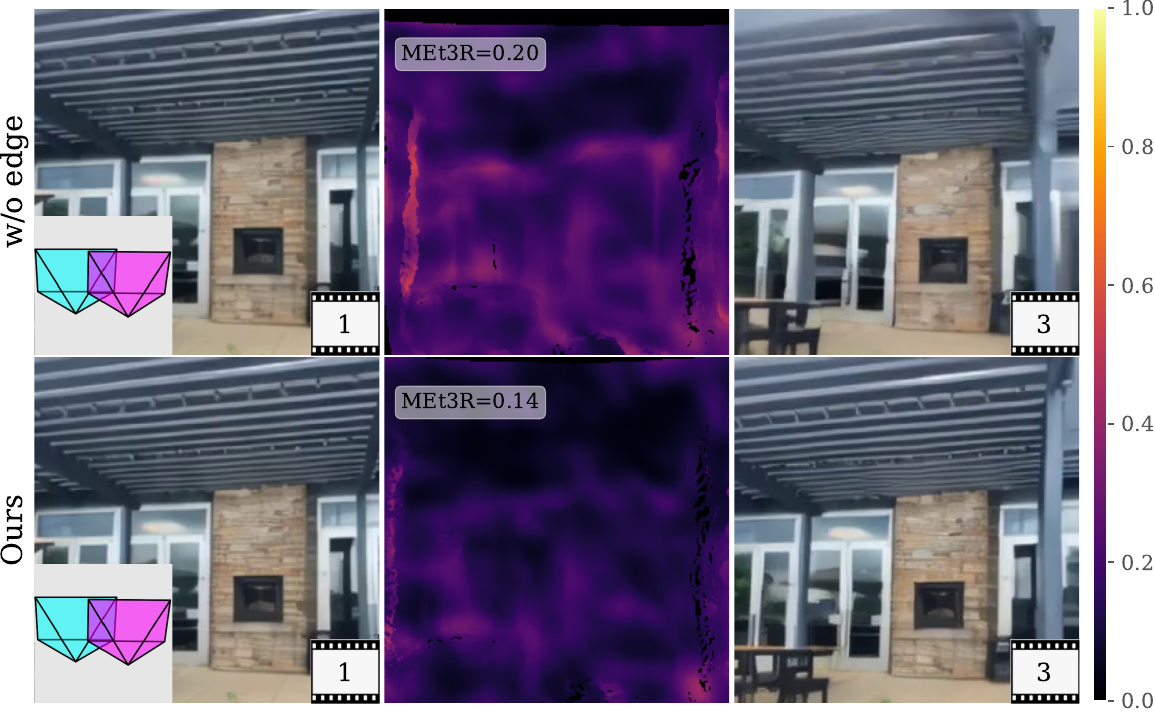}
    \caption{\textbf{Visual comparison of removing edge sampling}. 
    Over-sampling keypoints for $\mathcal{L}_\text{3DC}$ on edges during training leads to fewer artifacts on lines and corners, as can be seen from the MEt3R reprojection error map (middle column).}
    \label{fig:ablation_edge}
\end{figure}

\noindent\textbf{Edge-aware sampling.} When training \name, we bias the sampling of keypoints towards edges in images, as described in the main paper.
This design choice is ablated quantitatively in the main paper, and here we provide a qualitative result in \cref{fig:ablation_edge} illustrating its effectiveness.
We observe  that edge-aware sampling results in reduced artifacts on edges, which is also illustrated by the MEt3R~\cite{asim2025met3r} reprojection error map.

\section{Additional implementation details}\label{supp:impl}

\subsection{Analysis of 3D consistency experiments}
\label{supp:impl_analysis}
\noindent\textbf{Datasets.} For the main-paper correlation analysis of camera-controlled video diffusion models, we use RE10K~\cite{zhou2018stereo} and DL3DV~\cite{ling2024dl3dv} as evaluation datasets.
RE10K contains high-quality indoor and outdoor scenes and diverse camera trajectories~\cite{he2025cameractrl}. 
As the camera poses in RE10K are mostly smooth, we also report performance on DL3DV~\cite{ling2024dl3dv} which contains more outdoor scenes with more complex poses. 
Both are commonly used for camera-controlled video generation. 
We randomly sampled 300 test videos from each for our analysis.

\begin{table*}[t]
\centering
\caption{\textbf{Details of the pre-trained VDMs analyzed for 3D consistency.} The last column specifies the depth/optical flow models used during inference of each model.}
\vspace{-5pt}
\resizebox{\linewidth}{!}{
\begin{tabular}{m{2.5cm} m{12.5cm} m{3.5cm}}
\toprule
\textbf{Model}  &  \textbf{Version / Checkpoint} & \textbf{External Models}  \\ 
\midrule
MVGenMaster~\cite{cao2025mvgenmaster} & \url{https://github.com/ewrfcas/MVGenMaster} & DepthPro~\cite{bochkovskii2024depth}\\
MotionCtrl~\cite{wang2024motionctrl} & \url{https://github.com/TencentARC/MotionCtrl/tree/svd} (CMCM only) & N/A \\
CameraCtrl~\cite{he2025cameractrl} & \url{https://github.com/hehao13/CameraCtrl/tree/svd} & N/A \\
GWTF~\cite{burgert2025go} & \url{https://github.com/Eyeline-Labs/Go-with-the-Flow} & MoGe~\cite{wang2025moge}, RAFT~\cite{teed2020raft}\\
DaS~\cite{gu2025diffusion} & \url{https://github.com/IGL-HKUST/DiffusionAsShader} & MoGe~\cite{wang2025moge}\\
GEN3C~\cite{ren2025gen3c} & \url{https://github.com/nv-tlabs/GEN3C} & MoGe~\cite{wang2025moge} \\
RealCamI2V~\cite{li2025realcam} & \url{https://github.com/ZGCTroy/RealCam-I2V} & DepthAnythingV2~\cite{depth_anything_v2}\\
\bottomrule
\end{tabular}
}
\label{tab:analysis_models}
\end{table*}

\noindent\textbf{Models.} \cref{tab:analysis_models} summarizes the versions of the models used in our analysis. 
When running each model, we follow the official evaluation code and configurations. 
GEN3C, GWTF, MVGenMaster, DaS, and RealCamI2V require predicting a point cloud from the conditioning input image using an off-the-shelf depth estimation model, and projecting the point cloud to the desired target views. 
For GWTF, optical flow needs to be extracted from the re-projected frames to warp the input to the diffusion model.
We use the external models specified in the original papers or official implementations of the VDMs in our analysis, and note the versions in \cref{tab:analysis_models}. 
Predicting point clouds using a depth model and projecting the point cloud to the desired views requires that the depth prediction and input camera translations are aligned in scale.
Therefore, we also run the alignment methods specified in the original papers of the evaluated VDMs to obtain alignment parameters.  
For GWTF we use the same alignment parameters as GEN3C, as the former does not specify an alignment procedure.
We refer the readers to the original papers for more details.

The length of videos output by different models can vary from 14 (CameraCtrl) to 121 (GEN3C) frames. 
During inference, these VDMs require the same number of input camera poses as the number of frames to generate.
Since a 14-frame VDM cannot generate more frames, but it is possible to obtain fewer frames from a VDM that generates more than 14 frames, we limit the evaluated videos and diffusion representations to 14 frames.
That is, each test instance consists of an input image and a camera trajectory of length 14.
To evaluate a VDM that natively generates more frames than the available sequence length (\eg GEN3C), we pad the input camera trajectory by repeating the final camera pose in the sequence. At inference time, we retain only the generated frames/diffusion representations corresponding to the first 14 camera poses. Qualitatively we observe no noticeable impact from such padding on the generation quality of the initial 14 frames.

\noindent\textbf{Geometric Correspondence Evaluation.} To evaluate the multi-view consistency of diffusion representations within each VDM, we evaluate their internal features on the keypoint-free correspondence task~\cite{sun2021loftr} defined in the main-paper correlation analysis.
To extract features from the VDMs, a small amount of noise, which is equivalent to the 1/1000 of the respective highest noise levels, was added to the clean video latents from the VAE encoders.
We performed a layer search on each model in an attempt to find the most view-consistent feature encoding layer within each diffusion model.
Specifically, we followed \cite{el2024probing} and divided the models into four blocks equally, evaluating the features at the end of each block. 
The performance of the best-performing block was then reported.
In terms of the evaluation metric for geometric correspondence, considering the low resolution of the feature maps from VDMs (downsampled by a factor of eight from image resolution due to VAE encoding, with possibly additional downsampling within the diffusion model depending on the specific architecture), we calculate the pixel-space correspondence recall with a threshold of 50 pixels.

\noindent{\bf Clarification on GEN3C results.} In the main-paper correlation figure, GEN3C underperforms CameraCtrl on both benchmarks, whereas the original paper reports stronger performance~\cite{ren2025gen3c}.
The open-source GEN3C checkpoint evaluated here (specified in \cref{tab:analysis_models}) is Cosmos-based and trained at $\geq$720p, whereas the model in the original paper is based on Stable Video Diffusion and trained primarily on RE10K and DL3DV.
We hypothesize that CameraCtrl performs better in our evaluation because the gap between its training distribution and our test data is smaller, with both using RE10K at 576$\times$320.

\subsection{\name Intervention Training}
\label{supp:impl_vicodr} 
\noindent\textbf{Common intervention settings.}
For each base model, we construct a matched intervention pair: the baseline uses its original training objective, whereas the +\name variant adds $\mathcal{L}_\text{3DC}$.
Within each pair, we keep the training data, base-model trainable parameters, optimization settings, and number of training iterations identical.
At each iteration, $\mathcal{L}_\text{3DC}$ is enabled only when the sampled noise level lies within the first 20\% of the noising schedule, following~\cite{huang2025jog3r}.
We use $K=20{,}000$ keypoints, supervise features from the second-last layer of each base model following~\cite{jeong2025track4gen}, and implement $g_\phi$ as an eight-layer convolutional network.
We set the positive- and negative-sampling thresholds in $\mathcal{L}_\text{3DC}$ to $t_\text{pos}=0.005$ and $t_\text{neg}=0.1$.
Below, we report the loss weight $\lambda$ together with the remaining model-specific data and optimization details.

We evaluate all camera-controlled I2V interventions on the same held-out sets: 300 RE10K videos and a 1K-video subset of DL3DV~\cite{ling2024dl3dv}, which contains more challenging scenes and camera trajectories. 
For the camera-controlled T2V interventions, all MotionCtrl and UCPE variants are trained on RE10K using the captions provided by \cite{he2025cameractrl}.
Because text conditioning produces substantial stochastic variation in scene content, we evaluate every T2V variant using three inference seeds.

\noindent\textbf{CameraCtrl+\name.}
Both CameraCtrl and CameraCtrl+\name follow the official training setup~\cite{he2025cameractrl} and initialize the VDM from pre-trained SVD weights~\cite{blattmann2023stable}.
We use the original RE10K training set provided by CameraCtrl~\cite{he2025cameractrl}, comprising 14-frame videos at a resolution of 576$\times$320.
Both variants are trained with AdamW~\cite{loshchilov2019decoupled} at a learning rate of $3\times10^{-5}$.
Because the official batch size of 32 exceeds our hardware capacity, we use an effective batch size of eight through gradient accumulation over two backward passes.
CameraCtrl+JOG3R, CameraCtrl+GF, and CameraCtrl+Track4Gen use the same training setup.
All CameraCtrl variants are trained for 50K optimizer steps on two NVIDIA H200 GPUs.
For CameraCtrl+\name, we set $\lambda=0.2$ based on validation performance.

\noindent\textbf{DaS+\name.}
We initialize both variants from the released Diffusion-as-Shader checkpoint~\cite{gu2025diffusion}, which augments a CogVideoX-5B DiT with a tracking-map branch constructed from reprojected scene geometry (point maps).
We fine-tune DiT LoRA adapters with rank 128 and alpha 64 on RE10K while leaving the tracking-map conditioning pipeline unchanged.
Both variants are trained for 50K optimizer steps with AdamW, a learning rate of $10^{-4}$, and an effective batch size of eight.
We set $\lambda=0.2$.

\noindent\textbf{RealCamI2V+\name.}
We initialize both variants from the released RealCamI2V checkpoint~\cite{li2025realcam}, which combines a CogVideoX-5B I2V backbone~\cite{yang2024cogvideox} with ControlNet-XS camera conditioning and a geometry-rendered preview.
Following the same matched LoRA protocol, we fine-tune DiT LoRA adapters with rank 128 and alpha 64 on RE10K while retaining the released camera-conditioning components.
Both variants are trained for 50K optimizer steps with AdamW, a learning rate of $10^{-4}$, and an effective batch size of eight.
We set $\lambda=0.2$.

\noindent\textbf{Wan-CameraCtrl 1.3B+\name.}
We initialize both variants from the Wan-CameraCtrl 1.3B checkpoint~\cite{wan2025wan} (Wan2.1-Fun-V1.1-1.3B-Control-Camera) and fine-tune only DiT LoRA adapters, with rank 128 and alpha 64, on RE10K.
Both variants are trained for 50K optimizer steps with AdamW, a learning rate of $10^{-4}$, and an effective batch size of eight.
We set $\lambda=0.2$.

\noindent\textbf{MotionCtrl+\name.}
We use the AnimateDiff-based version of MotionCtrl~\cite{guo2023animatediff} for text-to-video generation.
Both variants are initialized from AnimateDiff and trained for 50K optimizer steps with AdamW, a learning rate of $3\times10^{-5}$, and a batch size of eight.
We set $\lambda=0.7$ based on validation performance.

\noindent\textbf{UCPE+\name.}
We initialize both variants from UCPE's released camera adapter on Wan2.1-T2V-1.3B~\cite{zhang2026unified,wan2025wan} and fine-tune only UCPE's camera-attention modules.
Both variants are trained for 50K optimizer steps with AdamW, a learning rate of $10^{-4}$, and an effective batch size of eight.
We set $\lambda=0.2$.

\noindent\textbf{ViFiGen+\name.}
In accordance with the original training protocol~\cite{xie2025mvgbench}, both ViFiGen and ViFiGen+\name are fine-tuned from the pre-trained SV3D backbone~\cite{voleti2024sv3d} using a curated 100K subset of ``kiui objects'', obtained from the original Objaverse dataset via the LGM filtering pipeline~\cite{tang2024lgm}.
This reduced 100K subset (compared to the original 150K used in ViFiGen) reflects our computational constraints.
As in ViFiGen, each object is rendered into 84 views, and then 21 views are randomly sampled per iteration using dynamic-orbit perturbations around a fixed camera ring. 
We use the official ViFiGen optimization setup with a reduced batch size: AdamW with a learning rate of $2\times10^{-5}$, a batch size of four, and 100K optimizer steps.
We set $\lambda=0.2$ based on validation performance.

\subsection{Evaluation metrics} 

\noindent\textbf{Camera-controlled I2V and T2V.}
For I2V, we assess image/video quality using PSNR and FVD~\cite{unterthiner2018towards}; for T2V, we report FVD.
To evaluate the \textit{3D consistency} of generated videos, we adopted two established metrics: MEt3R~\cite{asim2025met3r} and RPE~\cite{duan2025worldscore}. 
Both measure reprojection error after reconstructing a generated video in 3D: RPE uses GLOMAP~\cite{pan2024glomap} and measures error in pixel space, whereas MEt3R uses MASt3R~\cite{leroy2024grounding} and measures error in feature space.
The intuition behind both of these metrics is that a more 3D-consistent video will result in more accurate 3D reconstruction and keypoint matching, hence leading to smaller reprojection errors. 
Following~\cite{he2025cameractrl,li2025realcam}, we evaluate \textit{camera controllability} using rotation error (RErr) and translation error (TErr), which compare camera poses reconstructed from the generated video with COLMAP~\cite{schoenberger2016sfm} against the input trajectory.
The use of COLMAP or GLOMAP for each metric follows its original implementation.

\begin{table}[t]
\centering
\caption{\textbf{RErr and TErr results of \name on image-to-video generation.} Results are reported only on  successfully reconstructed videos common to all methods.}
\vspace{-5pt}
\label{tab:rot_trans_err_common_cameractrl}
\resizebox{0.8\linewidth}{!}{
\begin{tabular}{lcccc}
\toprule
                & \multicolumn{2}{c}{RE10K}                   & \multicolumn{2}{c}{DL3DV}                   \\ 
\cmidrule(l{5pt}r{5pt}){2-3} \cmidrule(l{5pt}r{5pt}){4-5}
\textbf{Model}  & \textbf{\textcolor{camMetric}{RErr} ($\downarrow$)} & \textbf{\textcolor{camMetric}{TErr} ($\downarrow$)} & \textbf{\textcolor{camMetric}{RErr} ($\downarrow$)} & \textbf{\textcolor{camMetric}{TErr} ($\downarrow$)} \\
\midrule
CameraCtrl      & 0.381               & 1.260                 & \textbf{0.725}               & 1.403                 \\
+ Track4Gen     & 0.498               & 1.945                 & 1.120               & 1.878                 \\
+ GF            & 0.385               & 1.322                 & 0.811               & 1.634                 \\
+ JOG3R         & 0.538               & 1.786                 & 1.080               & 1.761                 \\
    + \name & \textbf{0.302}               & \textbf{1.247}                 & 0.773               & \textbf{1.367}                 \\
\bottomrule
\end{tabular}
}
\end{table}

\begin{table}[t]
\centering
\caption{\textbf{RErr and TErr results of \name on text-to-video generation.}  Results are reported only on  successfully reconstructed videos common to all methods. Mean and standard deviation over three seeds are reported on RE10K.}
\vspace{-5pt}
\label{tab:rot_trans_err_common_motionctrl}
\resizebox{0.6\linewidth}{!}{
\begin{tabular}{@{}lcc@{}}
    \toprule
    \textbf{Model} & \textbf{\textcolor{camMetric}{RErr} ($\downarrow$)} & \textbf{\textcolor{camMetric}{TErr} ($\downarrow$)} \\
    \midrule
    MotionCtrl & 0.718 (0.017) & 3.509 (0.121) \\
    + \name   & 0.965 (0.039) & 4.917 (0.064) \\
    \bottomrule
\end{tabular}
}
\end{table}

Since SfM can fail on low-quality videos, or videos with insufficient camera baselines (which are rare), we apply a penalty value for RErr and TErr to penalize failed reconstructions and additionally report the reconstruction success rate (Succ \%) of each method.
For quantitative comparison on each test set, the penalty value for a metric (\ie RErr or TErr) was set to the 99th percentile of all scores obtained by all models for that metric on that test set.
For the main-paper I2V intervention results, the penalty values are 3.3 (RE10K) and 4.3 (DL3DV) for RErr, and 8.0 (RE10K) and 10.0 (DL3DV) for TErr.
For the main-paper T2V intervention results, the penalty values are 4.4 for RErr and 13.8 for TErr.
In \cref{tab:rot_trans_err_common_cameractrl} and \cref{tab:rot_trans_err_common_motionctrl} we additionally report RErr and TErr considering only common successfully reconstructed sequences between tested models.
These common-sequence results are consistent with the corresponding task-level comparisons in the main paper.

\noindent\textbf{Multi-view generation.}
We evaluate on the 6K-object OmniObject3D subset~\cite{wu2023omniobject3d} used by SV3D~\cite{voleti2024sv3d}.
Following MVGBench~\cite{xie2025mvgbench}, we fit two 3D Gaussian Splatting models to disjoint subsets of generated views.
We compute Chamfer distance and rendered depth error between the reconstructions to measure geometric consistency, and cPSNR, cSSIM, and cLPIPS between their rendered images to assess multi-view appearance consistency.

\end{document}